\documentclass[sigconf]{acmart}
\AtBeginDocument{%
  }

\setcopyright{acmlicensed}
\copyrightyear{2025}
\acmYear{2025}
\acmDOI{10.1145/3746027.3755545}
\acmConference[MM '25] {Proceedings of the 33rd ACM International Conference on Multimedia}{October 27--31, 2025}{Dublin, Ireland.}
\acmISBN{979-8-4007-2035-2/2025/10}
\PassOptionsToPackage{table}{xcolor}
\usepackage{microtype}
\usepackage{graphicx}
\usepackage{subfigure}
\usepackage{booktabs} 
\usepackage[table]{xcolor}
\usepackage{multirow}
\usepackage{colortbl}
\usepackage{multicol}
\usepackage{float}
\usepackage{algorithm}
\usepackage{algorithmic}

\begin{document}

\title{Analytic Continual Test-Time Adaptation for Multi-Modality Corruption}

\author{Yufei Zhang}
\orcid{0009-0009-3863-204X}
\affiliation{%
  \institution{South China University of Technology}
  \state{Guangzhou}
  \country{China}
}
\email{yufeizhang280@gmail.com}

\author{Yicheng Xu}
\affiliation{%
  \institution{Institute of Science Tokyo}
  \city{Tokyo}
  \country{Japan}}
\email{yxu040@e.ntu.edu.sg}
\orcid{0000-0003-2975-1206}

\author{Hongxin Wei}
\affiliation{%
  \institution{Southern University of Science and Technology}
  \city{Shenzhen}
  \country{China}}
  \email{weihx@sustech.edu.cn}
  \orcid{0000-0002-8973-2843}

\author{Zhiping Lin}
\affiliation{%
 \institution{Nanyang Technological University}
 \country{Singapore}}
 \email{EZPLin@ntu.edu.sg}
 \orcid{0000-0002-1587-1226}

\author{Xiaofeng Zou}
\affiliation{%
  \institution{South China University of Technology}
  \state{Guangzhou}
  \country{China}}
  \email{zouxiaofeng@scut.edu.cn}
  \orcid{0000-0002-5823-6345}

\author{Cen Chen}
\affiliation{%
  \institution{South China University of Technology}
  \state{Guangzhou}
  \country{China}}
  \email{chencen@scut.edu.cn}
  \orcid{0000-0003-1389-0148}

\author{Huiping Zhuang}
\affiliation{%
  \institution{South China University of Technology}
  \state{Guangzhou}
  \country{China}}
\email{hpzhuang@scut.edu.cn}
\orcid{0000-0002-4612-5445}

\renewcommand{\shortauthors}{Zhang et al.}

\begin{abstract}
  Test-Time Adaptation (TTA) enables pre-trained models to bridge the gap between source and target datasets using unlabeled test data, addressing domain shifts caused by corruptions like weather changes, noise, or sensor malfunctions in test time. Multi-Modal Continual Test-Time Adaptation (MM-CTTA), as an extension of standard TTA, further allows models to handle multi-modal inputs and adapt to continuously evolving target domains. However, MM-CTTA faces critical challenges such as catastrophic forgetting and reliability bias, which are rarely addressed effectively under multi-modal corruption scenarios. In this paper, we propose a novel approach, Multi-modality Dynamic Analytic Adapter (MDAA), to tackle MM-CTTA tasks. MDAA introduces analytic learning—a closed-form training technique—through Analytic Classifiers (ACs) to mitigate catastrophic forgetting. Furthermore, we design the Dynamic Late Fusion Mechanism (DLFM) to dynamically select and integrate reliable information from different modalities. Extensive experiments show that MDAA achieves state-of-the-art performance across the proposed tasks.
\end{abstract}



\received{9 April 2025}
\received[revised]{12 Jun 2025}
\received[accepted]{6 July 2025}

\maketitle

\section{Introduction}
Test-Time Adaptation (TTA) aims to enable pre-trained models to bridge the gap between the source domain and the target domain \citep{wangtent, liang2024comprehensive}. Unlike Unsupervised Domain Adaptation (UDA) \citep{zhang2015deep}, TTA performs adaptation without any source data (\textit{i.e.,} pre-trained dataset), which not only saves computational resources by avoiding retraining but also preserves the privacy of the source data. One key application of TTA is to address the problem of domain shift from source data to corrupted test data, where the corruption is often caused by weather changes, ambient noise or sensor malfunctions.  In these scenarios, Multi-Modal Test-Time Adaptation (MM-TTA) offers a distinct advantage over mono-modal TTA, as domain shifts often affect only a subset of modalities, while others remain relatively stable and informative \citep{radford2021learning, yang2024test}. This allows the model to leverage uncorrupted modalities to guide adaptation, thereby enhancing robustness in complex real-world environments.

\begin{figure*}
	\centering
        \setlength{\abovecaptionskip}{0.5pt}
	\centerline{\includegraphics[width=0.9\linewidth]{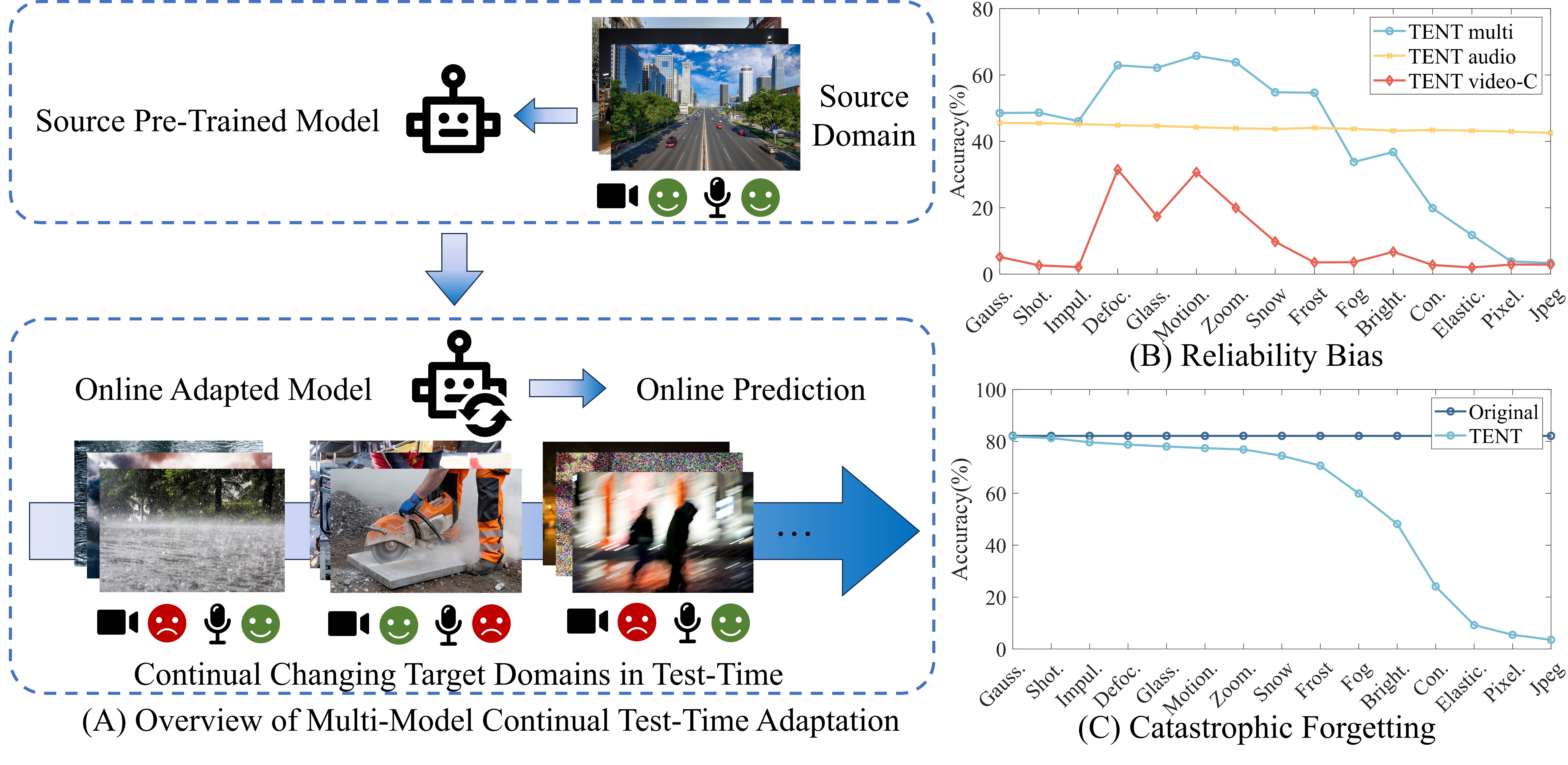}}
	\caption{\textbf{(A)} Overview of MM-CTTA.The pre-trained model is initially optimized on source data. During the test-time adaptation phase, however, it must continuously adapt to corrupted data arising from diverse real-world conditions (e.g., weather variations, ambient noise, or sensor malfunctions). (B-C) Illustration of key challenges in MM-CTTA, using the representative TTA method TENT \citep{wangtent} with CAV-MAE \citep{gong2022contrastive} as the backbone. \textbf{(B)} \textbf{Reliability bias}: As the dominant modality (video) becomes increasingly corrupted, the performance of the multi-modal network deteriorates, even falling below that of the audio-only network. \textbf{(C)} \textbf{Catastrophic forgetting}: The model's performance on the source data significantly declines during continuous adaptation. The ``Original" refers to the model's performance on the clean test set of Kinetics50 without any adaptation, while ``TENT" represents the performance on the same dataset after adaptation to the corresponding corruption. }
        \Description{}
        \label{fig.1}
\end{figure*}

To better reflect real-world settings where distribution shifts are continuous and multi-faceted, Multi-Modal Continual Test-Time Adaptation (MM-CTTA) has been proposed \citep{cao2023multi}. MM-CTTA addresses more complicated domain shifts, where the target distribution evolves both over time and across modalities. This setting is common in real-world scenarios, particularly for general-purpose AI agents (e.g., autonomous driving, robotic systems) that rely on multiple sensors for perception, as illustrated in Figure~\ref{fig.1}.A. However, achieving reliable MM-CTTA remains challenging, mainly due to two key issues. The first is \textbf{reliability bias}, which arises from intra-modality domain shifts and exacerbates inconsistencies within downstream fusion networks \citep{yang2024test}. The second is \textbf{catastrophic forgetting}, where knowledge acquired from the source domain deteriorates during continual adaptation, ultimately reducing the model’s generalization capability \citep{mccloskey1989catastrophic}.

MM-CTTA models must continuously adapt to multi-modal domain shifts while mitigating modality collapse and ensuring reliability in real-world scenarios. While recent researches on TTA have achieving promising results of in \textbf{catastrophic forgetting}~\citep{ wang2022continual, gao2023back, niu2022efficient} and \textbf{reliability bias}~\citep{yang2024test, lei2024two} respectively, directly combining CTTA methods with MM-TTA works often falls short. This is because dynamically changing corruption patterns magnify \textbf{reliability bias} and create additional complexities, as corruption can unpredictably switch between modalities during continual adaptation. Methods explicitly designed for MM-CTTA remain scarce. The closest existing approach, CoMAC \citep{cao2023multi}, focuses primarily on segmentation tasks in dynamic environments rather than addressing modality corruption.

In this paper, we propose a new approach named \textbf{M}ulti-modality \textbf{D}ynamic \textbf{A}nalytic \textbf{A}daptor (MDAA) to address the challenges in MM-CTTA. MDAA mainly comprises two primary components: (i) the Analytic Classifiers (ACs), and (ii) the Dynamic Late Fusion Mechanism (DLFM). The ACs update the model by addressing a recursive ridge regression problem, optimizing on both new target data and learned knowledge to avoid \textbf{catastrophic forgetting}, such approach have already been used in continual learning (CL) field~\citep{zhuang2022acil,zhuang2023gkeal}. Moreover, each feature type in the multi-modal model is configured with an individual AC in our framework. As a result, MDAA can derive predictions not only from the downstream fusion network but also from the feature encoders, marking a significant departure from conventional multi-modal paradigms.

To effectively utilize outputs from different ACs, we proposed a tailored late fusion strategy, DLFM, which compensates for and integrates predictions across modalities. DLFM selects reliable predictions from modality-specific ACs and propagates them to guide the updates of ACs associated with corrupted modalities, thereby alleviating \textbf{reliability bias}. To rigorously assess the robustness and adaptability of MDAA, we design two challenging evaluation settings: (1) adaptation over extremely long phases and (2) scenarios with alternating modality corruption. These settings provide a rigorous test of MDAA's robustness and adaptability under real-world conditions. The key contributions of this work are listed as follows: 

1).~We introduce ACs to TTA into the TTA framework to prevent the model from \textbf{catastrophic forgetting}. We propose DLFM to adaptively fuse features from multiple modalities, thereby mitigating \textbf{reliability bias}. 

2).~We design two MM-CTTA benchmarks featuring modality corruption switching and prolonged adaptation, better capturing real-world dynamics, pushing the boundaries of MM-CTTA.

3).~MDAA achieves state-of-the-art performance, outperforming prior methods by up to 6.22\% and 6.84\% in two benchmarks, while being more efficient in runtime and memory usage.
\section{Related Works}
\subsection{Test-Time Adaptation}

TTA focuses on enabling a pre-trained model to adapt to a new target domain without requiring access to the source domain data. However, simply relying on incorrect predictions during adaptation can mislead the update process, potentially leading to model collapse \citep{chen2019progressive}. TENT, as one of the first methods solve such issue, addresses this by only updating the model’s batch normalization (BN) layers through entropy minimization \citep{wangtent}. Subsequent research \citep{niutowards, gong2022note, pmlr-v202-zhou23e} has further mitigated this issue by filtering out low-confidence predictions using carefully designed thresholds and updating the layer normalization (LN) layers for more robust performance.

\begin{figure*}[htbp]
\centering
\includegraphics[width=0.71\linewidth]{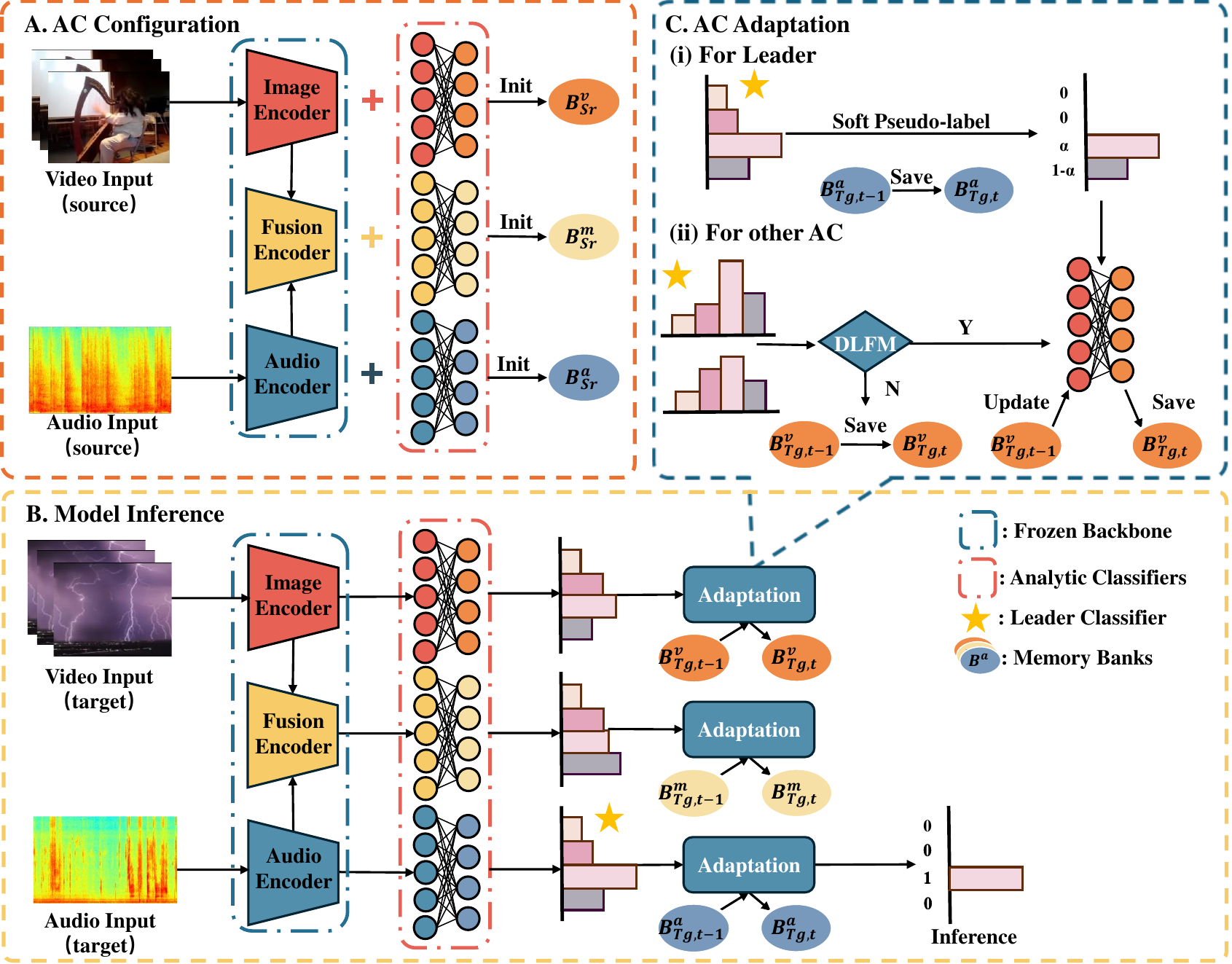}
\caption{The overview of MDAA. (A) Construct Analytic Classifiers (ACs) and initialize the memory bank $B\sim \left \{ \textbf{P}, \textbf{Q} \right \}$ to preserve knowledge from source datasets, using Eq.~\ref{eq7}. (B) In the inference pipeline, the classifier with highest MAP is selected as \textit{leader}. After adaptation, the model generates final predictions and updates the memory bank using Eq.\ref{eq12}. (C) During adaptation, ACs determine whether to update by comparing with the soft pseudo-label from the \textit{leader} through the Dynamic Late Fusion Mechanism (DLFM). The figure illustrates the process using the top 2 highest-probability labels.}
	\label{fig.2}
    \Description{}
\end{figure*}

\subsection{Multi-modality Test-Time Adaptation}

MM-TTA~\citep{zhao2025attention,guo2025smoothing,yang2024test,wang2020makes} seeks to enhance model reliability by incorporating multi-modal data into the TTA task. However, imbalances in inter-modal reliability can result in significant performance degradation, known as reliability bias \citep{wang2020makes}, shown in Figure~\ref{fig.1}.B. Existing MM-TTA models address this issue by independently updating BN or LN layers of each feature encoder, followed by a weighted fusion mechanisms \citep{shin2022mm, cao2023multi,xiong2024modality}. Although this allows more reliable modalities to carry more weight during fusion, the approach remains relatively shallow in terms of information integration.

A recent model called READ \citep{yang2024test} introduced a more advanced approach by fusing features through a Vision Transformer (ViT) \citep{dosovitskiy2020image} block, which allows for the preservation of parameters inherited from source data while effectively integrating inter-modal information \citep{vaswani2017attention, gong2022contrastive}. READ achieves reliable adaptation by modulating only the fusion layer within the attention module of the ViT block. Follow-up work \citep{lei2024two} aimed to improve performance by updating both the feature encoders and the fusion layer. However, this approach requires prior knowledge about which modalities are corrupted, making it less capable for real-world scenario.

\subsection{Analytic Continual Learning} \label{2.2}
When extending TTA to a continual setting, a significant challenge that emerges is catastrophic forgetting—a phenomenon where adaptation to the target domain leads to degradation of knowledge learned from the source domain and compromises generalization (as illustrated in Figure~\ref{fig.1}.C). Recent works have explored the integration of Continual Learning (CL) into TTA, demonstrating promising results \citep{niu2022efficient, cao2023multi}. Following this line of research, we incorporate a novel CL paradigm, Analytic Continual Learning (ACL), into our proposed MDAA framework.

ACL provides a global optimal solution through matrix inverse operations \citep{guo2004pseudoinverse}. To address the out-of-memory issue caused by large inverse matrices, \cite{zhuang2021blockwise} demonstrated that iterative computation using block-wise data achieves results equivalent to joint computation, making analytic learning highly effective in continual learning. By treating data from different time periods as blocks, ACL allows for recursive computation, with the final result being as accurate as if all data were processed simultaneously. Thanks to its non-forgetting properties, 
ACL has shown strong performance across various CL tasks in recent years \citep{zhuang2022acil, zhuang2023gkeal}.

\section{Multi-modality Dynamic Analytic Adapter}

In this section, we first formalize the MM-CTTA problem and introduce the key notations in Sec.\ref{3.1}. Then present the proposed MDAA framework, which incorporates ACs into pre-trained multi-modal encoders (Sec.\ref{3.2}). Subsequently, we detail the AC adaptation in Sec.\ref{3.3} and introduce DLFM for updating process in Sec.\ref{3.4}. An overview of MDAA is provided in Figure~\ref{fig.2}.

\subsection{Problem Definition and Notations}\label{3.1}

To illustrate our method, we consider an audio-video classification task, which generalizes to other multi-modal scenarios without loss of generality. In MM-CTTA, the weight in pre-trained models $\textbf{W}_\textup{Sr}$ is trained on a labeled source dataset ${D}_{\textup{Sr}}\sim \left\{{\textbf{X}}_{\textup{Sr}}^{a},{\textbf{X}}_{\textup{Sr}}^{v},{\textbf{Y}}_{\textup{Sr}}\right\}$ in source domain $\textup{Sr}$, where ${\textbf{X}}_{\textup{Sr}}^{a}$ and ${\textbf{X}}_{\textup{Sr}}^{v}$ represent the audio and video training data respectively. ${\textbf{Y}}_{\textup{Sr}}$ represents the corresponding one-hot label set. During the adaptation phase, for each timestamp $t$ in the target domain $\textup{Tg}$, the model can only perform inference and update itself based on the unlabeled test dataset ${D}_{\textup{Tg},t}\sim\left\{{\textbf{X}}_{\textup{Tg},t}^{a},{\textbf{X}}_{\textup{Tg},t}^{v}\right\}$. The target domain varies with $t$ and the whole adaptation process can be represented as $\textbf{W}_ {\textup{Sr}} \to \textbf{W}_ {\textup{Tg},1} \to\dots\to \textbf{W}_ {\textup{Tg},t}$.

\subsection{Source Model and Analytic Classifier Configuration}\label{3.2}

The multi-modal backbone follows a typical extraction–fusion architecture, where training data from different modalities are first processed by modality-specific feature encoders and subsequently fused through a fusion encoder. During this process, features from different encoders are projected into a unified higher-dimensional space to enhance representational capacity, and are then passed through a linear classifier to produce prediction labels, as illustrated in Figure~\ref{fig.2}.A. These linear classifiers are referred to as Analytic Classifiers (ACs). During the adaptation phase, both the feature encoders and the projection layers are kept frozen, while only the ACs are updated. Since features from different modalities are processed by ACs in a unified manner, we omit modality-specific distinctions for simplicity.

The weights ${\textbf{W}}_{\textup{Sr}}$ for the ACs are optimized via ridge regression, initialized using a least squares objective, a strategy commonly employed in classification tasks~\cite{zhuang2022acil, fang2024air, zhuang2023gkeal, yang2008stagewise}. To address the issue of class imbalance, we reformulate the optimization objective as follows:
\begin{align}\label{eq3}
\underset{{\textbf{W}}_{\textup{Sr}}}{\text{argmin}} {\textstyle \sum_{k=1}^{N_{\textup{Sr}}}} \omega_{k}   \left\lVert{\textbf{y}}_{\textup{Sr},k} - {\textbf{x}}_{\textup{exf},k}{\textbf{W}}_{\textup{Sr}}\right\rVert_{\text{F}}^{2}+ {\gamma} \left\lVert{\textbf{W}}_{\textup{Sr}}\right\rVert_{\text{F}}^{2},
\end{align}
where $\left\lVert\cdot\right\lVert_{\text{F}}$ indicates the Frobenius norm, ${\gamma}$ is the regularization parameter and $N_{\textup{Sr}}$ is the samples size of $D_{\textup{Sr}}$. $\omega_{k}$, ${\textbf{x}}_{\textup{exf},k}$ and ${\textbf{y}}_{\textup{Sr},k}$ represents the weight, expanded feature vector and one-hot label of sample k in ${D}_{\textup{Sr}}$, while the weight is further defined as 
\begin{align}\label{eq4}
\omega_{k}= \frac{N_\textup{Sr}}{N_{C}\times N_{c\mid k}},
\end{align}
where $N_{C}$ is the number of classes in $D_\textup{Sr}$ and $N_{c\mid k}$ is the number of samples in $D_\textup{Sr}$ from category $c$ to which sample $k$ belongs. Following the ridge regression solution, the solution to optimization problem is given as
\begin{align} \label{eq5}
\notag
 \hat{\textbf{W}}_{\textup{Sr}} & =( {\textstyle \sum_{k=1}^{N_{\textup{Sr}} }} {\tilde{\textbf{x}} }_{\textup{exf},k}^\top\tilde{\textbf{x}}_{\textup{exf},k}+\gamma \mathbf{I})^{-1}{\textstyle \sum_{k=1}^{N_{\textup{Sr}} }}{\tilde{\textbf{x}} }_{\textup{exf},k}^\top{\tilde{\textbf{y}}}_{\textup{Sr},k}\\
 &=( {\tilde{\textbf{X}} }_{\textup{exf},\textup{Sr}}^\top\tilde{\textbf{X}}_{\textup{exf},\textup{Sr}}+\gamma{\textbf{\textbf{I}}})^{-1}{\textstyle }{\tilde{\textbf{X}} }_{\textup{exf},\textup{Sr}}^\top{\tilde{\textbf{Y}}}_{\textup{Sr}},
\end{align}
where $\tilde{\textbf{x}}_{\textup{exf},k} = \sqrt{\omega_{k}} {\textbf{x}}_{\textup{exf},k}$ and ${\tilde{\textbf{y}} }_{\textup{Sr},k} = \sqrt{\omega_{k}}{\textbf{y}}_{\textup{Sr},k}$. In addition to the classifier weights $\hat{\textbf{W}}_{\textup{Sr}}$, a memory bank $B_{\textup{Sr}}$ needs to be constructed during the training phase. Unlike other methods \citep{cao2023multi, zhang2023adanpc, xiong2024modality}, our memory bank contains only two types of matrices, which can be represented as $B_{\textup{Sr}}\sim \left \{ \textbf{P}_{\textup{Sr}}, \textbf{Q}_{\textup{Sr}} \right \}$, where 
\begin{align}
\begin{cases}
  \textbf{P}_{\textup{Sr}}= {\tilde{\textbf{X}} }_{\textup{exf},\textup{Sr}}^\top\tilde{\textbf{X}}_{\textup{exf},\textup{Sr}}+\gamma{\textbf{I}},\\
  \textbf{Q}_{\textup{Sr}}={\tilde{\textbf{X}}}_{\textup{exf},\textup{Sr}}^\top{\tilde{\textbf{Y}}}_{\textup{Sr}}.
\label{eq7}
\end{cases}
\end{align}

Both $\textbf{P}_{\textup{Sr}}$ and $\textbf{Q}_{\textup{Sr}}$ are used to extract and preserve the learned knowledge from the source dataset, which cannot be accessed during adaptation. Therefore $\hat{\textbf{W}}_{\textup{Sr}}$ can be further rewritten as 
\begin{align}
\hat{\textbf{W}}_{\textup{Sr}} = \textbf{P}_{\textup{Sr}}^{-1}\textbf{Q}_{\textup{Sr}}.
\end{align}

\subsection{Model Inference and Analytic Classifier Adaptation}\label{3.3} 

During test time, MDAA determines the final prediction for each sample by selecting the output distribution with the highest maximum a posteriori (MAP) score across all ACs, as higher predictive probabilities generally reflect greater confidence~\cite{boudiaf2022parameter}. Subsequently, DLFM is employed to further filter the samples used for model updates, which will be elaborated upon in the following section.

Since ACL performs global optimization, it considers not only the currently selected input sample ${\textbf{X}}_{\textup{Tg},t}$ at timestamp $t$, but also all previously observed data, including ${\textbf{X}}_{\textup{Sr},k}$ and ${\textbf{X}}_{\textup{Tg},1:t-1}$. The corresponding optimization problem for the weight $\bar{\textbf{W}}_{\textup{Tg},t}$ is formulated as:
\begin{align} \label{eq10}
\notag
\underset{{\textbf{W}}_{\textup{Tg},t}}{\text{argmin}}& {\textstyle \sum_{k=1}^{N_{\textup{Sr}}}} \omega_{k} \left\lVert{\textbf{y}}_{\textup{Sr},k} - {\textbf{x}}_{\textup{exf},k}{\textbf{W}}_{\textup{Tg},t}\right\rVert_{\text{F}}^{2} +\\
& \left\lVert{\bar {\textbf{Y}}}_{\textup{Tg},1:t} - {\textbf{X}}_{\textup{exf},1:t}{\textbf{W}}_{\textup{Tg},t}\right\rVert_{\text{F}}^{2} +{\gamma} \left\lVert{\textbf{W}}_{\textup{Tg},t}\right\rVert_{\text{F}}^{2},
\end{align}
where $\bar{\textbf{Y}}$ denotes for soft pseudo-labels of filtered samples. We adopt soft labels for AC updates as they retain a richer representation of uncertainty compared to hard labels \citep{muller2019does, hinton2015distilling}. For each filtered sample, we sort and choose its top-$n$ class positions $C = {c_1, c_2, \dots, c_n}$ and assign weights $\alpha_1, \alpha_2, \dots, \alpha_n$ $(\sum_{i=1}^{n} \alpha_i = 1)$. In this paper, we define the weights as $\alpha_{i} = round({(n+1-i)}/{\sum_{i=1}^{n}i})$, and construct the soft pseudo-label accordingly as:
\begin{align} \label{SPS}
    \bar{\textbf{y}}_i =\left\{\begin{matrix}
    \alpha _{i} &, i\in C \\
    0 &, \text{otherwise.}  \\
    \end{matrix}\right.
\end{align}
Such reconsrtuction is proven to be useful, as shown in ablation study in subsection~\ref{4.3}.

Due to the definition of TTA, all datasets (\textit{i.e.,} ${\textbf{X}}_{\textup{Sr},k}$ and ${\textbf{X}}_{\textup{Tg},1:t-1}$) prior to timestamp $t$ are not accessible when solving the optimization problem. However, with the aid of the memory bank $B_{\textup{Tg},t}\sim \left \{ \textbf{P}_{\textup{Tg},t}, \textbf{Q}_{\textup{Tg},t} \right \}$, the solution can still be computed, as stated follow.

\textbf{Theorem 1.} The optimal solution to Formula \ref{eq10} is given as
\begin{align}\label{eq11}
\hat{\textbf{W}}_{\textup{Tg},t}&=\textbf{P}_{\textup{Tg},t}^{-1}\textbf{Q}_{\textup{Tg},t},
\end{align}
where
\begin{align}
\left\{\begin{matrix}
 \textbf{P}_{\textup{Tg},t}=\textbf{P}_{\textup{Tg},t-1}+ \textbf{X}_{\textup{exf},t}^\top \textbf{X}_{\textup{exf},t} \label{eq12},\\
\textbf{Q}_{\textup{Tg},t}=\textbf{Q}_{\textup{Tg},t-1}+ \textbf{X}_{\textup{exf},t}^\top {\bar{\textbf{Y}}}_{\textup{Tg},t}.
\end{matrix}\right.
\end{align}

\textit{Proof} of \textbf{Theorem 1} is provided in Appendix~\ref{A}. It can be seen that, only features and labels that are accessible in phase $t$ are involved in the adaptation, which fits the definition of TTA. Solution in Eq.~\ref{eq11} is the global optimal solution for all the samples that have appeared, therefore it is not affected by batch size and timestamp, thus solving the catastrophic forgetting at the root. 


\subsection{Dynamic Late Fusion Mechanism}\label{3.4}

 
Although AC-based adaptation effectively mitigates the forgetting issue, it is worth noting that employing the least squares loss as the objective function may be less robust to outliers compared to the cross-entropy loss. This vulnerability increases the risk of model collapse when adapting with corrupted modality inputs. This raises a critical question:\textbf{ how can we reliably select trustworthy samples for updating each AC?}

\begin{figure}
	\centering
        \setlength{\abovecaptionskip}{1pt} 
	\includegraphics[width=1\linewidth]{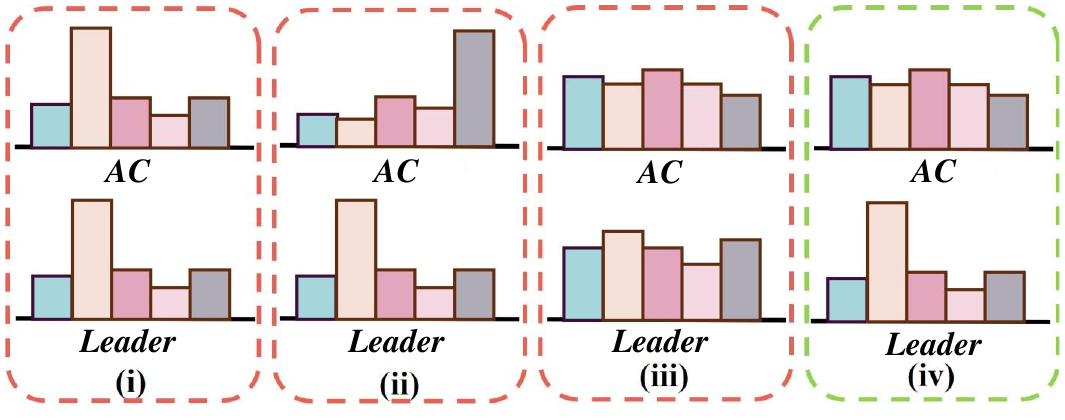}
	\caption{Four possible relationship between the probability distribution of \textit{Leader} and compared AC. Only samples belong to case (iv) will be used to update AC.}
        \Description{}
	\label{fig.3}
\end{figure}

To tackle this challenge, we propose a novel fusion strategy termed Dynamic Late Fusion Mechanism (DLFM). For each input sample, DLFM first identifies the AC yielding the highest MAP score, designating it as the \textit{Leader}. During adaptation, each AC is updated independently based on its MAP score relative to the \textit{Leader}. All possible relationships and update rules can be enumerated as four scenarios (see Figure~\ref{fig.3}).

(i). \textbf{Close Distributions}: When the MAP scores of the AC and the \textit{Leader} are nearly identical and correspond to the same label, the AC remains unchanged to avoid reinforcing existing knowledge and causing class imbalance.

(ii). \textbf{Different Labels with Close Distributions}: If the MAP scores are similar but correspond to different labels, no update is made to avoid introducing potential errors due to uncertainty in the \textit{Leader}'s prediction.

(iii). \textbf{Evenly Distributed Probabilities}: If both the AC and the \textit{Leader} show evenly distributed probabilities with no clear preference, the AC is not updated to prevent adjustments based on uncertain predictions.

(iv). \textbf{Significant Difference}: If the \textit{Leader}'s MAP is significantly higher than the AC's, the AC is updated to reflect the \textit{Leader}'s confident prediction.

To summarize, cases (i),(ii) and (iii) all belong to a small gap between MAP scores of AC and \textit{Leader}, while case (iv) belongs to a larger gap. Therefore, given a pre-defined threshold $\theta$, Rules of DLFM can be noted as:
\begin{align} \label{DLFM}
\left\{\begin{matrix}
\text{Accept} ,& Leader_{MAP}-AC_{MAP}\geqslant \theta \\
\text{Reject} ,& \text{otherwise.} \\
\end{matrix}\right.
\end{align}

DLFM leverages the complementary and consistent information across modalities to enhance adaptation. For complementarity, it selects the most reliable prediction—typically from an uncorrupted modality—as a pseudo label to update other classifiers, aligning corrupted inputs with reliable targets to handle domain shifts. For consistency, when all classifiers predict the same label, indicating high confidence and low likelihood of corruption, no update is performed to avoid unnecessary adaptation and prevent class imbalance. Since our recursive formulation already preserves prior knowledge, additional updates are unnecessary. By updating each classifier individually, DLFM maximizes modality-specific information while reducing reliability bias.

It is also important to note that in Eq.~\ref{eq10} we treat source data and target data separately, we adjust the category balance for the source dataset by assigning weights to each class. In contrast, it is challenging to apply this approach to the target dataset, as it is impossible to know the exact number of samples in each category within the target domain. However, with the help of the DLFM, category balance can still be maintained, as unimportant samples have been filtered out. Given that the weights added to the sample data are expected to average out to 1, the impact of each sample from both the source and target domains on the model can be considered equivalent.
\begin{table*}[ht]
\caption{Comparison with SOTA methods on \textbf{audio} \textbf{progressive single-modality corruption} task in terms of classification Top-1 accuracy (\%), using dataset \textbf{Kinetics50-C} and \textbf{VGGSound-C} in severity level 5. The best results for each domain are highlighted in \textbf{bold}. $^*$ means we revise the method from BN to LN for fair comparison.}
	\centering
	\resizebox{0.81\linewidth}{!}{
        \renewcommand\arraystretch{1}
		\begin{tabular}{|c|c|cccccc|c|cccccc|c|}
			\hline 
			\multirow{2}*{Method}&\multirow{2}*{Type}&\multicolumn{7}{c|}{Kinetics50-C}&\multicolumn{7}{c|}{VGGSound-C}\\ 
                \cline{3-16}
			&& Gauss. & Traff. & Crowd & Rain & Thund. & Wind & Avg. & Gauss. & Traff. & Crowd & Rain & Thund. & Wind & Avg. \\ \hline 
			&&\multicolumn{6}{c|}{$t\quad\xrightarrow[\quad\quad\quad\quad\quad\quad\quad\quad\quad\quad\quad\quad\quad\quad]{}$}&&\multicolumn{6}{c|}{$t\quad\xrightarrow[\quad\quad\quad\quad\quad\quad\quad\quad\quad\quad\quad\quad\quad\quad]{}$}& \\ 
                \hline 
                Source& - &\textbf{73.97} & 65.17 & 67.88 & 70.24 & 68.00 & 70.44 & 69.28 & 37.29 & 21.24 & 16.89 & 21.81 & 27.36 & 25.66 & 25.04 \\
                TENT*& TTA &73.02 & 63.36 & 45.31 & 37.02 & 34.57 & 34.01 & 47.88 & 0.68 & 0.28 & 0.28 & 0.28 & 0.28 & 0.28 & 0.35 \\
                SAR& TTA &72.18 & 70.36 & 48.30 & 37.67 & 36.21 & 39.09 & 50.64 & 16.09 & 4.50 & 4.33 & 3.60 & 12.00 & 5.51 & 7.67 \\
                EATA& CTTA &73.91 & 65.29 & 68.24& 70.51 & 68.28 & 70.48 & 69.45 & \textbf{40.39} & 31.99 & 31.91 & 32.38 & \textbf{39.24} & 33.95 & 34.98 \\
                MMTTA*& MM-TTA &17.03 & 1.99 & 1.99 & 1.99 & 1.99 & 1.99 & 4.50 &0.41&0.33&0.33&0.33&0.33&0.33&0.34 \\
                READ& MM-TTA &65.65 &49.34&46.45&44.94&44.45 &43.72 &49.09 &18.73 &8.52 &8.01 &5.90 &8.30 &	4.82 &9.05 \\
                READ+SRM& MM-CTTA&52.23 &38.30 &40.68&46.61&47.02 &48.60 &45.57 &18.36 &7.88 &8.34 &8.01&	16.26&10.21 &11.51 \\ 
                \rowcolor{lightgray!45}
                MDAA& MM-CTTA &72.87 & \textbf{71.45} & \textbf{72.91} & \textbf{72.26} & \textbf{73.20} & \textbf{73.80} & \textbf{72.75} & 38.80&\textbf{34.91}&\textbf{34.63} &\textbf{34.59} &37.70&\textbf{35.85}&\textbf{36.08} \\\hline
                
                &&\multicolumn{6}{c|}{$t\quad\xleftarrow[\quad\quad\quad\quad\quad\quad\quad\quad\quad\quad\quad\quad\quad\quad]{}$}&&\multicolumn{6}{c|}{$t\quad\xleftarrow[\quad\quad\quad\quad\quad\quad\quad\quad\quad\quad\quad\quad\quad\quad]{}$} &\\ 
                \hline 
                TENT*& TTA &43.27 & 42.96 & 43.81 & 60.19 & 69.17 & 70.17 & 54.93 & 0.30 & 0.30 & 0.30 & 0.30 & 0.30 & 0.39 & 0.32 \\
                SAR& TTA &41.81 & 27.94 & 24.41 & 40.47 & 42.90 & 70.36 & 41.32 & 14.91 & 4.56 & 4.61 & 3.72 & 12.44 & 5.94 & 7.70 \\
                EATA& CTTA &73.91 & 65.32 & 68.18 & 70.49 & 68.26 & 70.47 & 69.44 & \textbf{40.22} & 33.69 & 31.61 & 32.64 & \textbf{39.67} & 32.81 &  35.11 \\
                MMTTA*& MM-TTA &1.99 & 1.99 & 1.99 & 1.99 & 1.99 & 39.21 & 8.19 & 0.33 & 0.30 &	0.31 & 0.30 & 0.61 &1.49 &0.56 \\
                READ& MM-TTA &48.95 &48.70&48.72&51.99 &62.69 &65.34 &54.40 &8.99&6.00&7.71 &7.71 &11.56 &	11.89 &8.98 \\
                READ+SRM& MM-CTTA &48.59 &47.39 &46.48 &41.11 &38.01 &51.90&	45.58& 18.50 &	7.93 &	8.36 &7.96 &	16.28 &10.11 &11.53\\
                \rowcolor{lightgray!45}
                MDAA& MM-CTTA &\textbf{74.86} & \textbf{72.63} & \textbf{72.87} & \textbf{72.26} & \textbf{73.4} & \textbf{72.02} & \textbf{73.01} & 38.95 &\textbf{35.46}&\textbf{34.66}&\textbf{34.70}&37.31&\textbf{35.20}&\textbf{36.05} \\ \hline 
		\end{tabular}
}
\label{tab1}
\end{table*}
\begin{table*}[ht]
\caption{Comparison with SOTA methods on \textbf{video} \textbf{progressive single-modality corruption} task in terms of classification Top-1 accuracy (\%), with dataset \textbf{Kinetics50-C} in severity level 5.}
	\centering
	\resizebox{0.95\linewidth}{!}{
        \renewcommand\arraystretch{1}
		\begin{tabular}{|c|c|ccccccccccccccc|c|}
			\hline 
			Method&Type& Gauss. & Shot. & Impul. & Defoc. & Glass. & Motion. & Zoom. & Snow & Frost & Fog & Bright. & Cont.& Elastic. &Pixel. &Jpeg & Avg. \\ 
                \hline 
			&&\multicolumn{15}{c|}{$t\quad\xrightarrow[\qquad\qquad\qquad\qquad\qquad\qquad\qquad\qquad\qquad\qquad\qquad\qquad\qquad\qquad]{}$}& \\ 
                \hline 
                Source &-&48.74 & 49.80 & 48.99 & \textbf{67.68}& 61.84 & \textbf{70.88} & \textbf{66.18} & 61.35 & 61.39 & 45.34 & 75.95 & 51.87 & 65.77 & 68.78 & 66.10 & 60.71 \\
                TENT*&TTA&16.23 & 2.07 & 2.03 & 2.08 & 2.06 & 2.03 & 2.03 & 2.03 & 2.03 & 2.03 & 2.03 & 2.03 & 2.03 & 2.03 & 2.03 & 2.98 \\
                SAR&TTA&38.36 & 35.97 & 34.51 & 44.40 & 48.86 & 50.77 & 47.53 & 43.59 & 35.81 & 42.54 & 52.11 & 35.44 & 50.20 & 40.15 & 50.73 & 43.40 \\
                EATA&CTTA&48.80 & 49.82 & 49.03 & 67.66 & 61.98 & 70.84 & 66.16 & 61.64 & 61.54 & 45.40 & \textbf{75.99} & 51.95 & 65.88 & 68.71 & 66.08 & 60.77 \\
                MMTTA*&MM-TTA&14.31 & 2.64 & 2.03 & 2.03 & 2.03 & 2.03 & 2.03 & 2.03 & 2.03 & 2.03 & 2.03 & 2.03 & 2.03 & 2.03& 2.03 & 2.89 \\
                READ&MM-TTA&12.76 &2.03 &2.02&3.27 &2.60 &2.64 &2.64 &2.50 &2.39 &2.42 &2.44 &2.40&2.42 &	2.40 &2.41 &3.16 \\
                READ+SRM&MM-CTTA &41.90 &42.29&42.27 &54.78 &54.37 &58.21 &56.70 &53.70 &52.68 &46.12 &61.97 &	45.16 &58.63 &55.32 &54.54 &52.68 \\
                \rowcolor{lightgray!45}
                MDAA&MM-CTTA&\textbf{54.89} & \textbf{55.25} & \textbf{55.32} & 63.89 & \textbf{62.49} & 67.26 & 65.86 & \textbf{64.32} & \textbf{65.31} & \textbf{61.86} & 73.20 & \textbf{61.60} & \textbf{67.83} & \textbf{69.22} & \textbf{68.69} & \textbf{63.80} \\
                \hline
                &&\multicolumn{15}{c|}{$t\quad\xleftarrow[\qquad\qquad\qquad\qquad\qquad\qquad\qquad\qquad\qquad\qquad\qquad\qquad\qquad\qquad]{}$}& \\ 
                \hline
                TENT*&TTA&2.03 & 2.03 & 2.03 & 2.03 & 2.03 & 2.03 & 2.03 & 2.03 & 2.03 & 2.03 & 2.03 & 2.03 & 2.10 & 3.47 & 53.26 & 5.55 \\
                SAR&TTA&34.75 & 35.08 & 35.89 & 42.70 & 45.99 & 49.43 & 50.12 & 44.08 & 42.42 & 40.02 & 57.54 & 35.56 & 48.86 & 57.22 & \textbf{66.26} & 45.73 \\
                EATA&CTTA&48.81 & 49.79 & 49.02 & 67.71 & 61.96 & \textbf{70.88} & 66.17 & 61.56 & 61.51 & 45.38 & \textbf{75.96} & 51.90 & 65.90 & \textbf{68.76} & 66.09 & 60.76 \\
                MMTTA*&MM-TTA&1.99 & 1.99 & 1.99 & 1.99 & 1.99 & 1.99 & 1.99 & 1.99 & 1.99 & 1.99 & 1.99 & 1.99 & 1.99 & 1.99 & 23.88 & 3.45 \\
                READ&MM-TTA&2.03 &2.03 &2.03 &2.03&2.03 &2.03 &2.03 &2.03&2.09 &2.03 &2.22&2.04 &2.43 &3.65& 	24.34 &3.67\\
                READ+SRM&MM-CTTA&41.91&42.42&42.46&54.97&54.32&58.30&56.83&53.66&52.93&46.03&62.05&45.00&58.69 	&55.20&54.60&53.19\\
                \rowcolor{lightgray!45}
                MDAA&MM-CTTA&\textbf{67.32} & \textbf{67.48} & \textbf{67.76} & \textbf{68.98} &\textbf{ 67.60} & 69.59 & \textbf{68.49} & \textbf{66.46} &\textbf{ 66.18} & \textbf{63.14} & 72.87 & \textbf{59.33} & \textbf{66.59} & 67.64 & 65.25 & \textbf{66.98} \\ \hline 
		\end{tabular}
}
\label{tab2}
\end{table*}
\begin{table*}[ht]
\caption{Comparison with SOTA methods on \textbf{video} \textbf{progressive single-modality corruption} task in terms of classification Top-1 accuracy (\%), with dataset \textbf{VGGSound-C} in severity level 5.}
	\centering
	\resizebox{0.95\linewidth}{!}{
        \renewcommand\arraystretch{1}
		\begin{tabular}{|c|c|ccccccccccccccc|c|}
			\hline 
			Method&Type& Gauss. & Shot. & Impul. & Defoc. & Glass. & Motion. & Zoom. & Snow & Frost & Fog & Bright. & Cont.& Elastic. &Pixel. &Jpeg & Avg. \\ 
                \hline 
			&&\multicolumn{15}{c|}{$t\quad\xrightarrow[\qquad\qquad\qquad\qquad\qquad\qquad\qquad\qquad\qquad\qquad\qquad\qquad\qquad\qquad]{}$}& \\ 
                \hline 
                Source &-&53.02 & 52.90 & 52.98 & \textbf{57.20 }& \textbf{57.38} & 58.37 & 57.48 & 56.40 & 56.46 & 55.41 & \textbf{59.16} & 53.73 & 57.22 & 56.44 & 57.33 & 56.10 \\
                TENT*&TTA&51.48 & 50.70 & 50.87 & 51.15 & 50.90 & 51.09 & 50.82 & 50.65 & 50.75 & 50.73 & 50.73 & 50.58 & 50.70 & 50.73 & 50.70 & 50.84 \\
                SAR&TTA&43.12 & 38.99 & 37.77 & 42.43 & 43.84 & 43.61 & 43.79 & 42.13 & 41.26 & 42.83 & 43.84 & 39.34 & 42.75 & 43.49 & 40.38 & 41.97 \\
                EATA&CTTA&53.57 & 53.70 & 53.57 & 57.00 & 57.29 & \textbf{58.46} & \textbf{57.77} & 56.24 & \textbf{56.57} & 55.45 & 59.06 & 54.13 & \textbf{58.24} & \textbf{57.22} & 57.38 & 56.38 \\
                MMTTA*&MM-TTA&0.46 & 0.34 & 0.34 & 0.34 & 0.34 & 0.34 & 0.34 & 0.34 & 0.34 & 0.34 & 0.34 & 0.34 & 0.34 & 0.34 & 0.34 & 0.35 \\
                READ&MM-TTA&33.12&10.41&4.05&0.77&0.37&0.35&0.32&0.29&0.29&0.29&0.29&0.29&0.29&0.29&0.29&3.45 \\
                READ+SRM&MM-CTTA&41.33&41.37&41.32&45.50&45.68&47.18&46.61&44.66&44.88&42.97&47.07&40.01&46.18 	&45.41&45.24&44.36\\
                \rowcolor{lightgray!45}
                MDAA&MM-CTTA&\textbf{55.13} & \textbf{55.29} & \textbf{55.30} & 56.91 & 57.20 & 57.78 & 57.32 & \textbf{56.52} & 56.25 & \textbf{56.14} & 58.11 & \textbf{55.32} & 57.06 & 56.27 & \textbf{57.39} & \textbf{56.53}\\
                \hline
                &&\multicolumn{15}{c|}{$t\quad\xleftarrow[\qquad\qquad\qquad\qquad\qquad\qquad\qquad\qquad\qquad\qquad\qquad\qquad\qquad\qquad]{}$}& \\ 
                \hline
                TENT*&TTA&52.68 & 52.74 & 52.49 & 53.45 & 53.45 & 53.50 & 53.10 & 53.35 & 53.67 & 52.83 & 55.86 & 51.82 & 56.81 & \textbf{57.46} & 57.43 & 54.04 \\
                SAR&TTA&40.31 & 39.24 & 38.33 & 41.53 & 41.36 & 44.43 & 43.59 & 42.46 & 41.11 & 41.52 & 42.24 & 38.97 & 43.49 & 40.97 & 46.22 & 41.72 \\
                EATA&CTTA&53.63 & 53.60 & 53.61 & 57.06 & 57.27 & \textbf{58.35} & \textbf{57.83} & 56.22 & \textbf{56.74} & 55.73 & \textbf{59.16} & 54.09 & \textbf{58.19} & 57.27 & 57.35 & 56.41 \\
                MMTTA*&MM-TTA&0.34 & 0.34 & 0.34 & 0.34 & 0.34 & 0.34 & 0.34 & 0.34 & 0.34 & 0.34 & 0.34 & 0.34 & 0.34 & 0.34 & 8.32 & 0.87 \\
                READ&MM-TTA&13.26 &13.39 &13.38 &13.66 &13.72 &13.91 &13.92 &14.25 &14.84 &18.15 &20.68 &22.85 	 &29.46 &40.99 &50.38 &20.46 \\
                READ+SRM&MM-CTTA&41.46&41.25&41.41&45.77&45.95&47.17&46.85&44.73&45.03&43.37&46.90&39.93&46.32 	&45.46&45.18&44.45\\
                \rowcolor{lightgray!45}
                MDAA&MM-CTTA&\textbf{55.30} & \textbf{55.38} & \textbf{55.25} & 56.90 & 57.19 & 57.79 & 57.32 & \textbf{56.50} & 56.31 & \textbf{56.22} & 58.13 &\textbf{ 55.28} & 57.07 & 56.30 & \textbf{57.47} & \textbf{56.56} \\ \hline 
		\end{tabular}
            }
\label{tab3}
\end{table*}
\begin{table*}[ht]
\caption{Comparison with SOTA methods on \textbf{interleaved modality corruption} task in terms of classification Top-1 accuracy (\%), with dataset \textbf{Kinetics50-C} in severity level 5. A-C and V-C indicates the corrupted modality in current phase.}
	\centering
	\resizebox{0.95\textwidth}{!}{
        \renewcommand\arraystretch{1}
		\begin{tabular}{|c|cc|c|cc|c|cc|c|ccc|c|cc|c|cc|c|cc|c|}
			\hline 
			\multirow{2}*{Method}&\multicolumn{2}{c|}{V-C}&{A-C}&\multicolumn{2}{c|}{V-C}&{A-C}&\multicolumn{2}{c|}{V-C}&{A-C}&\multicolumn{3}{c|}{V-C}&{A-C}&\multicolumn{2}{c|}{V-C}&{A-C}&\multicolumn{2}{c|}{V-C}&{A-C}&\multicolumn{2}{c|}{V-C}& \multirow{2}*{Avg.}\\
                \cline{2-22}
                & Gauss. & Shot. &Gauss. & Impul. & Defoc. & Traff. & Glass. & Motion.& Crowd & Zoom.  & Snow & Frost & Rain & Fog  & Bright. & Thund. & Cont.& Elastic. & Wind &Pixel. &Jpeg  & \\ 
                \hline 
			&\multicolumn{21}{c|}{$t\quad\xrightarrow[\qquad\qquad\qquad\qquad\qquad\qquad\qquad\qquad\qquad\qquad\qquad\qquad\qquad\qquad\qquad\qquad\qquad\qquad\qquad\qquad\qquad\qquad]{}$}& \\ \hline 
                Source &48.71 & 49.98 & 74.03 & 48.98 & 67.69 & 67.89 & 61.82 & 70.92 & 70.29 & 66.14 & 61.36 & 61.35 & 68.02 & 45.34 & 75.94 & 65.20 & 51.82 & 65.84 & 70.38 & 68.73 & 66.11 & 63.17 \\
                TENT* &48.77 & 48.34 & 74.11 & 46.38 & 62.83 & 67.22 & 62.30 & 68.35 & 69.32 & 64.61 & 54.64 & 57.36 & 66.32 & 46.22 & 63.95 & 37.12 & 38.87 & 42.77 & 40.62 & 10.13 & 5.40 & 51.22 \\  
                SAR &48.65 & 49.81 & \textbf{74.15} & 48.53 & 66.87 & 65.68 & 62.56 & 70.67 & 68.00 & 66.45 & 58.80 & 60.42 & 69.96 & 47.69 & 75.19 & 67.89 & 50.96 & 65.54 & 70.03 & 66.77 & 63.67 & 62.78 \\
                EATA &48.81 & 49.70 & 74.07 & 48.96 & 67.75 & 65.35 & 61.96 & \textbf{70.95} & 67.98 & 66.05 & 61.60 & 61.49 & 70.40 & 45.29 & \textbf{76.11} & 68.13 & 51.85 & 65.98 & 70.51 & 68.80 & 66.15 & 63.23 \\
                MMTTA*&48.63 & 49.20 & 56.24 & 47.79 & 47.52 & 4.76 & 42.96 & 29.87 & 1.96 & 4.13 & 2.27 & 1.96 & 1.96 & 1.96 & 1.92 & 2.03 & 1.96 & 1.92 & 1.96 & 1.92 & 1.99 & 16.90 \\
                READ &51.18 & 53.62 & 73.88 & 54.73 & \textbf{68.67} & 67.95 & \textbf{67.07} & 70.14 & 68.84 & \textbf{67.62} & 62.68 & 64.90 & 68.24 & 59.46 & 71.43 & 68.16 & 52.60 & 66.35 & 66.18 & 62.85 & 63.35 & 64.28 \\
                READ+SRM &46.18 &45.39&55.52&44.25& 56.45&44.43&55.42&59.41&49.61&57.63&54.45&53.65&50.92&50.60&62.39&52.52&45.93&59.55&52.69&55.92&56.06&52.81 \\
                \rowcolor{lightgray!45}
                MDAA &\textbf{55.04} & \textbf{55.91} & 73.64 & \textbf{55.89} & 63.78 & \textbf{73.12} & 63.54 & 67.62 & \textbf{74.90} & 67.00 & \textbf{65.60} & \textbf{67.21} & \textbf{75.44} & \textbf{64.89} & 72.69 & \textbf{76.94} & \textbf{65.45} &\textbf{ 69.21 }& \textbf{76.95} & \textbf{70.94} & \textbf{71.16} & \textbf{67.95} \\
                \hline
                &\multicolumn{21}{c|}{$t\quad\xleftarrow[\qquad\qquad\qquad\qquad\qquad\qquad\qquad\qquad\qquad\qquad\qquad\qquad\qquad\qquad\qquad\qquad\qquad\qquad\qquad\qquad\qquad\qquad]{}$} &\\ \hline 
			Source&48.73 & 49.75 & 74.02 & 48.99 & 67.61 & 65.21 & 61.93 & 70.87 & 67.87 & 66.17 & 61.36 & 61.43 & 70.27 & 45.32 & 75.88 & 67.97 & 51.84 & 65.74 & 70.47 & 68.74 & 66.10 & 63.16 \\
                TENT* &6.72 & 10.03 & 52.92 & 23.89 & 53.44 & 63.96 & 60.40 & 64.98 & 67.75 & 62.93 & 60.28 & 61.20 & 67.26 & 52.50 & 71.88 & 58.44 & 51.24 & 67.56 & 70.84 & 68.87 & 66.41 & 55.40 \\
                SAR &47.97 & 48.73 & 72.21 & 48.46 & 66.39 & 66.75 & 63.69 & 70.28 & 68.00 & 66.33 & 59.24 & 59.85 & 70.17 & 46.42 & 75.52 & 67.59 & 51.12 & 65.23 & 70.24 & 68.31 & 66.18 & 62.79 \\
                EATA &48.76 & 49.76 & 73.89 & 48.86 & 67.75 & 65.20 & 61.89 & 70.67 & 68.28 & 66.16 & 61.64 & 61.44 & 70.33 & 45.57 & \textbf{75.96} & 68.30 & 51.88 & 65.87 & 70.51 & 68.59 & 65.98 & 63.20 \\
                MMTTA* &1.96 & 1.96 & 2.00 & 1.96 & 1.96 & 2.00 & 2.00 & 2.03 & 1.96 & 2.03 & 1.96 & 2.00 & 1.96 & 2.52 & 18.31 & 4.88 & 43.11 & 49.71 & 62.28 & 55.69 & 60.04 & 15.35 \\
                READ &51.49 & 52.29 & 71.09 & 50.70 & 62.44 & 63.72 & 65.01 & 66.42 & 66.39 & 64.86 & 60.97 & 64.44 & 67.85 & 62.48 & 74.68 & 72.12 & 54.12 & 69.15 & 70.27 & \textbf{69.74} & \textbf{68.37} & 64.22 \\
                READ+SRM &44.57& 44.24 &55.61 &	44.36 &	56.04& 	44.20 &55.60 &59.01 &	50.48& 	57.75 &54.41 &	53.47 	&50.80 &50.95 &62.64 &52.76 &46.44 &60.00 &53.65 &56.84 &58.81&	52.98\\ 
                \rowcolor{lightgray!45}
                MDAA &\textbf{70.44} & \textbf{70.21} &\textbf{ 77.35} & \textbf{70.12} & \textbf{72.13} & \textbf{76.56} & \textbf{70.05} & \textbf{72.44} & \textbf{76.07} & \textbf{71.46} & \textbf{68.94} & \textbf{68.72} & \textbf{75.92} & \textbf{66.51} & 73.32 & \textbf{75.10} &\textbf{ 61.78} & \textbf{68.05} & \textbf{73.52} & 68.56 & 64.91 & \textbf{71.06} \\
 			\hline 
		\end{tabular}
}
\label{tab4}
\end{table*}
\begin{table*}[ht]
\caption{Comparison with SOTA methods on \textbf{interleaved modality corruption} task in terms of classification Top-1 accuracy (\%), with dataset \textbf{VGGSound-C} in severity level 5.}
	\centering
	\resizebox{0.95\textwidth}{!}{
        \renewcommand\arraystretch{1}
		\begin{tabular}{|c|cc|c|cc|c|cc|c|ccc|c|cc|c|cc|c|cc|c|}
			\hline 
			\multirow{2}*{Method}&\multicolumn{2}{c|}{V-C}&{A-C}&\multicolumn{2}{c|}{V-C}&{A-C}&\multicolumn{2}{c|}{V-C}&{A-C}&\multicolumn{3}{c|}{V-C}&{A-C}&\multicolumn{2}{c|}{V-C}&{A-C}&\multicolumn{2}{c|}{V-C}&{A-C}&\multicolumn{2}{c|}{V-C}& \multirow{2}*{Avg.}\\
                \cline{2-22}
                 & Gauss. & Shot. &Gauss. & Impul. & Defoc. & Traff. & Glass. & Motion.& Crowd & Zoom. & Snow & Frost & Rain & Fog & Bright. & Thund. & Cont.& Elastic. & Wind &Pixel. &Jpeg & \\ 
                \hline 
			&\multicolumn{21}{c|}{$t\quad\xrightarrow[\qquad\qquad\qquad\qquad\qquad\qquad\qquad\qquad\qquad\qquad\qquad\qquad\qquad\qquad\qquad\qquad\qquad\qquad\qquad\qquad\qquad\qquad]{}$}& \\ \hline 
                Source &53.05 & 52.91 & 37.32 & 52.98 & 57.19 & 21.24 & 57.37 & 58.37 & 16.89 & 57.45 & 56.37 & 56.47 & 21.82 & 55.41 & 59.19 & 27.37 & 53.75 & 57.19 & 25.66 & 56.44 & 57.33 & 47.23 \\
                TENT* &53.19 & 52.80 & 3.43 & 50.52 & 53.15 & 0.65 & 51.83 & 53.10 & 0.60 & 52.64 & 50.91 & 51.89 & 0.67 & 51.15 & 51.73 & 2.13 & 48.68 & 50.76 & 0.79 & 50.38 & 50.22 & 37.20 \\
                SAR&53.16 & 53.33 & 34.26 & 53.17 & 56.94 & 11.27 & 57.14 & 58.29 & 9.30 & 57.65 & 56.11 & 56.78 & 13.36 & 55.94 & 57.87 & 17.95 & 53.13 & 56.51 & 20.46 & 55.30 & 55.79 & 44.94 \\
                EATA &53.63 & 53.70 & \textbf{40.51} & 53.59 & 57.21 & 30.81 & 57.45 & 58.49 & 29.80 & 57.85 & 56.37 & 56.85 & 30.55 & 56.72 & 59.13 & \textbf{37.29} & 54.31 & 58.27 & 32.58 & 57.28 & 57.55 & 50.00 \\
                MMTTA*& 8.27 & 0.34 & 0.34 & 0.34 & 0.34 & 0.34 & 0.34 & 0.34 & 0.34 & 0.34 & 0.34 & 0.34 & 0.34 & 0.34 & 0.34 & 0.34 & 0.34 & 0.34 & 0.34 & 0.34 & 0.34 & 0.72 \\
                READ &53.78 & 53.91 & 39.83 & 54.17 & \textbf{57.81} & 26.00 & \textbf{58.14} & \textbf{59.42} & 21.63 & \textbf{59.03} & \textbf{57.38} & \textbf{58.29} & 22.79 & \textbf{57.71} & \textbf{59.32} & 26.07 & \textbf{55.46} & \textbf{58.36} & 18.29 & \textbf{57.36} & \textbf{57.63} & 48.21 \\
                READ+SRM &42.34& 	41.70&22.88&41.73&45.39&10.93&45.62&46.75&11.23&46.28&45.00&44.82&10.92&42.59&47.43&19.22&40.47&46.43&12.57&45.02&45.39&35.94 \\ 
                \rowcolor{lightgray!45}
                MDAA &\textbf{55.09} & \textbf{55.31} & 38.60 & \textbf{55.31} & 56.89 & \textbf{34.83} & 57.20 & 57.69 & \textbf{34.65} & 57.35 & 56.47 & 56.27 & \textbf{34.28} & 56.17 & 58.05 & 36.86 &55.33 & 57.00 & \textbf{35.53} & 56.28 & 57.35 & \textbf{50.60}\\\hline
                &\multicolumn{21}{c|}{$t\quad\xleftarrow[\qquad\qquad\qquad\qquad\qquad\qquad\qquad\qquad\qquad\qquad\qquad\qquad\qquad\qquad\qquad\qquad\qquad\qquad\qquad\qquad\qquad\qquad]{}$}& \\ \hline 
			Source&53.01 & 52.88 & 37.31 & 52.97 & 57.20 & 21.25 & 57.42 & 58.41 & 16.89 & 57.49 & 56.37 & 56.49 & 21.81 & 55.43 & 59.16 & 27.37 & 53.74 & 57.19 & 25.66 & 56.42 & 57.29 & 47.23 \\
                TENT* &24.80 & 42.44 & 1.65 & 49.78 & 51.65 & 0.32 & 51.64 & 51.52 & 0.34 & 51.35 & 50.51 & 50.89 & 0.43 & 51.27 & 52.07 & 1.14 & 51.14 & 54.94 & 1.64 & 56.50 & 56.92 & 35.85 \\
                SAR &51.90 & 51.78 & 25.90 & 51.30 & 54.73 & 5.11 & 54.82 & 56.52 & 7.82 & 56.31 & 54.63 & 55.49 & 13.44 & 55.12 & 57.59 & 15.07 & 53.57 & 56.92 & 15.04 & 56.53 & 57.30 & 43.19 \\
                EATA &53.77 & 53.65 & \textbf{40.39} & 53.61 & 57.17 & 30.49 & 57.36 & 58.59 & 30.16 & 57.82 & 56.21 & 56.64 & 31.22 & 56.77 & \textbf{59.24} & \textbf{37.44} & 54.30 & \textbf{58.18} & 33.01 & 57.46 & 57.53 & 50.05 \\
                MMTTA* &0.34 & 0.34 & 0.34 & 0.34 & 0.34 & 0.34 & 0.34 & 0.34 & 0.34 & 0.34 & 0.34 & 0.34 & 0.34 & 0.34 & 0.34 & 0.34 & 0.34 & 0.34 & 8.36 & 18.74 & 46.32 & 3.79 \\
                READ &54.35 & 54.56 & 25.03 & 54.38 & \textbf{57.99 }& 17.67 & \textbf{57.76} & \textbf{58.74} & 20.57 & \textbf{58.66} & \textbf{56.91} & \textbf{57.58} & 20.81 & \textbf{58.04} & 59.10 & 33.82 & \textbf{55.54} & 58.13 & 32.75 & \textbf{57.80} & \textbf{58.34} & 48.03 \\
                READ+SRM &41.84& 	41.72&21.72&41.78&45.17&10.47&45.62&46.73&11.84&46.34&44.98&44.83&10.98&42.69&47.49&19.50&40.59&46.53&12.48&44.97&45.96&35.92 \\ 
                \rowcolor{lightgray!45}
                MDAA &\textbf{55.30} & \textbf{55.41} & 38.64 & \textbf{55.29} & 56.91 & \textbf{35.40} & 57.14 & 57.81 & \textbf{34.85} & 57.34 & 56.48 & 56.29 & \textbf{34.52} & 56.23 & 58.24 & 37.22 &55.31 & 57.13 & \textbf{35.12} & 56.30 & 57.47 & \textbf{50.69} \\
 			\hline 
		\end{tabular}
}
\label{tab5}
\end{table*}

\section{Experiments}
In this section, we evaluate MDAA under two challenging settings which is detailed in Sec.~\ref{4.1}. Sec.~\ref{4.2} compares MDAA with other SOTA methods through extensive experiments and Sec.\ref{4.3} conducts ablation studies on MDAA’s components and computational efficiency. Implementation details are in Appendix~\ref{D}.

\subsection{Benchmarks and Settings}\label{4.1}
To evaluate the model performance, we designed two tasks specifically for MM-CTTA. The first task named \textbf{Progressive single-modality corruption}, sequentially corrupts one modality while keeping the other clean, assessing the model’s resistance to catastrophic forgetting. It follows an online setting, where the model processes one sample at a time, leading to an extreme long learning phase. The second task, called \textbf{interleaved modality corruption}, continually alternates corruption between the two modalities. While most methods perform poorly in the online setting due to severe catastrophic forgetting, this task uses a batch size of 64 during test time to emphasize assessing the model’s ability to adapt to dynamic reliability biases. We utilize two datasets for both tasks: Kinetics50 \citep{kay2017kinetics} and VGGSound \citep{chen2020vggsound}. Following \cite{yang2024test}, we construct their corrupted type named Kinetics50-C and VGGSound-C, each dataset includes 15 types of video corruptions and 6 audio corruptions at the largest severity (level 5).

\subsection{Performance Comparison}\label{4.2}
In this section, we reproduce various TTA methods under the MM-CTTA setting. These include typical TTA methods (TENT \citep{wangtent}, SAR \citep{niutowards}), CTTA methods (EATA \citep{niu2022efficient}), and MM-TTA methods (MMTTA \citep{shin2022mm}, READ \citep{xiong2024modality}). Additionally, we introduce an MM-CTTA variant, READ+SRM, which integrates READ with the stochastic restoration mechanism (SRM) from CoTTA~\cite{wang2022continual}. To ensure fairness, all methods use the same pre-trained model CAV-MAE \citep{gong2022contrastive} as the multi-modal encoder. For methods that update BN layers, we instead update LN layers to align with the ViT structure of encoder. We also evaluate the source model, a baseline trained only on the source dataset and kept frozen during testing.

The performances of various methods on \textbf{progressive single-modality corruption} are summarized in Tables~\ref{tab1}, \ref{tab2}, and \ref{tab3}. Table~\ref{tab1} reports results on audio corruption, while Tables~\ref{tab2} and \ref{tab3} focus on video corruption. Notably, most comparison methods collapse on VGGSound-C, performing worse than source models. In contrast, EATA achieves better results by limiting parameter updates, enabling more stable adaptation. MDAA outperforms prior methods by 3.00\%-3.57\% (audio) and 3.03\%-6.22\% (video) on Kinetics50-C, and by 0.94\%-1.10\% (audio) and 0.13\%-0.18\% (video) on VGGSound-C, on average. While MDAA exhibits suboptimal performance in scenarios where data corruption minimally impacts source model performance—attributable to its reduced parameter space for adaptation compared to baselines—it nevertheless demonstrates superior overall adaptability. MDAA consistently retains its advantage in later stages of the tasks, demonstrating superior robustness against catastrophic forgetting in MM-CTTA.

The comparison results of \textbf{interleaved modality corruption} tasks are shown in Table~\ref{tab4} and \ref{tab5}. In this task, EATA, which is more good at memorization, is not dominant in the task of highlighting reliability bias. READ, which is specifically designed to address intra-modal reliability bias, demonstrates strong performance in this area. However, its effectiveness is limited to video corruption in MM-CTTA, as performance drops significantly during audio corruption. In contrast, MDAA is well-adapted to the corruption of different modalities, outperforming READ by 2.39\%-6.84\% and EATA by 0.60\%-7.28\% on average across Kinetics50-C and VGGSound-C. 

\begin{figure}[t]
	\centering
	\includegraphics[width=0.95\linewidth]{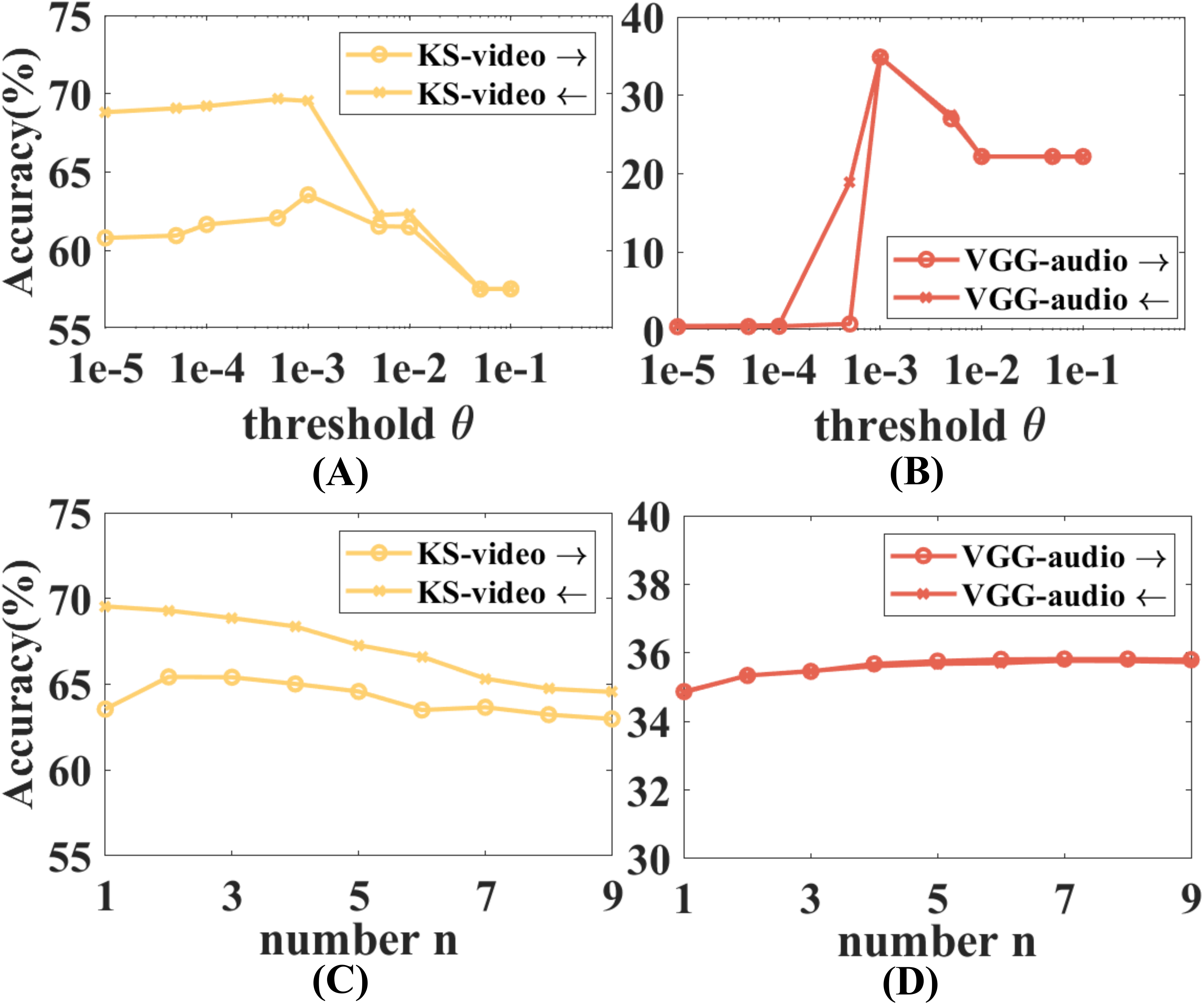}
	\caption{(A-B) Comparison of different thresholds $\theta$. (C-D) Comparison of reconstructed pseudo-labels with varying category numbers.}
	\label{fig.4}
        \Description{}
\end{figure}
\subsection{Ablation Study}\label{4.3}
All studies in this section use the video-corrupted Kinetics50-C (KS-video) and audio-corrupted VGGSound-C (VGG-audio) datasets at severity level 5 with a batch size of 64.

\textbf{Component analysis.} To verify the effectiveness of each MDAA component, we adopt an ablation study \textit{w.r.t} three components as shown in \autoref{Component}. As observed, due to the vulnerability of least square loss, the model using only ACs underperforms, with an average accuracy 0.5\%-0.57\% lower than READ (using cross-entropy loss) on KS-video. The model even collapses on VGG-audio, with an average accuracy of 0.47\%-0.56\%. The introduction of DLFM consistently improves model performance across all types of corruption, achieving the gains of 1.73\%-7.52\% on KS-video and 33.74\%-33.90\% on VGG-audio. Soft labeling introduces a small amount of uncertainty compared to hard labeling, it effectively alleviates the negative impact caused by incorrect pseudo labels during TTA. Although such uncertainty may lead to slightly lower performance in some tasks where corruptions in these tasks caused little effect on the model classification, soft labeling represents a favorable trade-off, particularly when dealing with highly complex datasets such as VGGSound. Furthermore, combining DLFM with soft labels yields the best overall performance.

\textbf{Reliable selection threshold.} To examine the effect of the threshold $\theta$ on the DLFM, we plot the model's performance with $\theta$ of from 1$e$-5 to 1$e$-1 in Figure~\ref{fig.4}(A-B). The performance of the model on both datasets exhibits an increasing and then decreasing trend. When $\theta$ is close to $0$, the ACs are updated for nearly every sample, introducing more error. Conversely, when $\theta$ increases too much, the ACs do not learn from new inputs since no samples can pass through the DLFM, leading to a decline in performance. Therefore, value of $\theta$ should not be too small, otherwise the model will not be able to filter out the update samples with low execution. For both datasets in the experiment, a reasonable value of $\theta$ is around 1$e$-3.

\begin{table}[ht]
 \caption{Ablation studies on different component combinations. Grey denotes the default setting.}
     \resizebox{0.85\columnwidth}{!}{
     \renewcommand\arraystretch{1}
     \begin{tabular}{|l|cc|cc|}
     \hline 
          \multirow{2}*{Method}&\multicolumn{2}{c|}{KS-video}&\multicolumn{2}{c|}{VGG-audio}  \\ \cline{2-5}
          &$\xrightarrow{}$&$\xleftarrow{}$&$\xrightarrow{}$&$\xleftarrow{}$\\ \hline
          READ&62.32 &62.59 &23.93 &22.39 \\
          MDAA (ACs+Hard)&61.82&62.02&0.47&0.56 \\
          MDAA (ACs+Soft)&61.33&67.47&32.50&33.00\\
          MDAA (ACs+DLFM+Hard)&63.55&\textbf{69.54}&34.87&34.85\\ 
          \rowcolor{lightgray!45}
          MDAA (ACs+DLFM+Soft)&\textbf{65.43}&69.30&\textbf{35.82}&\textbf{35.77}\\ \hline  
     \end{tabular}
     }
     \label{Component}
\end{table}
\begin{table}[ht]
 \caption{Comparison of runtime and VRAM usage.} \label{vram}
     \centering
     \resizebox{0.85\linewidth}{!}{
     \renewcommand\arraystretch{1}
     \begin{tabular}{|c|c|c|c|c|}
     \hline 
          \multirow{2}*{Method}&\multicolumn{2}{c|}{KS-video}&\multicolumn{2}{c|}{VGG-audio}  \\ \cline{2-5}
          &time(s)&memory(MB)&time(s)&memory(MB)\\ \hline
          TENT*&1085 &37620&2571 &37622 \\
          SAR &2149  &37620&4823&39220 \\
          EATA&889  &67286&2313&38300\\
          MMTTA*&2574  &71732&5966&68840\\
          READ&1174  &9176&2611&9178\\
          \rowcolor{lightgray!45}MDAA &1024 &8784&1682&8750\\ \hline  
     \end{tabular}
     }
\end{table}

\textbf{Soft label reconstruction.} Figure~\ref{fig.4}(C-D) shows the model results with $n$ ranging from $1$ to $9$. While using soft labels inevitably introduces error, there are cases such as KS-video $\xrightarrow{}$ where one-hot labels may outperform soft labels. However, using soft pseudo-labels remains beneficial as datasets become larger and more complex. The choice of $n$ needs to be moderate, as larger datasets with more categories tend to require larger $n$, while more error is likely to be introduced as $n$ increases. In this work, we determine to use top $2$ and $7$ classes for label reconstruction in Kinetics50-C and VGGSound-C respectively.

\textbf{Runtime and VRAM usage.} To demonstrate MDAA’s practicality, we compare its runtime and VRAM usage against various methods in Table~\ref{vram}. MDAA achieves superior efficiency with lower memory overhead and faster execution, thanks to its unique design. Unlike other methods, MDAA eliminates backpropagation, reducing gradient computation and optimizer storage costs. The primary computational cost of MDAA lies in matrix inversion, with a time complexity of $O(\varphi^3)$, where $\varphi$ is the expanded feature dimension, fixed at 8000 in our experiments. 

\section{Conclusion}
MM-CTTA shows great potential in real-world scenarios, as it enables pre-trained models to handle multi-modal inputs and adapt to continuously changing target domains. However, it also suffers from critical challenges such as reliability bias and catastrophic forgetting. To tackle these issues, we propose a new paradigm, MDAA, which introduces analytic learning into TTA. Instead of simply adapting the model to the target domain, MDAA integrates the target domain into source domain, and thus prevent network from forgetting. With the help of DLFM, model is able to dynamically and comprehensively process the information provided by each modality and use reliable samples to update. Moreover, compared to most SOTA TTA methods, MDAA achieves faster computational times and lower memory usage, highlighting its robust applicability in real-time deployment. In the future, we will try to adapt this method to more modalities to solve more challenging tasks.

\begin{acks}
This research was supported by the National Natural Science Foundation of China (Grant No. 62306117), the Guangzhou Basic and Applied Basic Research Foundation (2024A04J3681), and GJYC program of Guangzhou (2024D03J0005).
\end{acks}

\bibliographystyle{ACM-Reference-Format}
\bibliography{my_article}

\newpage
\appendix
\section{Pseudo-code for MDAA} \label{B}
The pseudo-code of MDAA is shown in Algorithm \ref{alg1}. For the pre-trained models, we integrate an individual AC for each network block, using the source dataset to initialize the classifiers as well as the memory bank. During the inference and adaptation periods, the model reconstructs the output labels for each sample using the soft pseudo-label and determines which ACs need to be updated through the DLFM.

\begin{algorithm} 
    \caption{\textbf{M}ulti-modality \textbf{D}ynamic \textbf{A}nalytic \textbf{A}daptor (MDAA)} 
	\label{alg1} 
	\begin{algorithmic}
		\REQUIRE Source datasets ${D}_{\textup{Sr}}\sim\{{\textbf{X}}_{\textup{Sr}}^{a},{\textbf{X}}_{\textup{Sr}}^{v},{\textbf{Y}}_{\textup{Sr}}\}$ and target datasets ${D}_{\textup{Tg},t}\sim\{{\textbf{X}}_{\textup{Tg},t}^{a},{\textbf{X}}_{\textup{Tg},t}^{v}\}$, pre-trained network $\Phi _{\textup{Sr}}$.
 \STATE 1. Training phase:
 \STATE (1) integrate AC for each network block;
 \STATE (2) Determine the parameters of each AC using ${D}_{\textup{Sr}}$ through Eq.~\ref{eq5};
 \STATE (3) Initialize the memory bank $B_S$ through Eq.~\ref{eq7}.
 \STATE 2. Inference and Adaptation phase:
 \FOR{ Samples in each batch}
 \STATE (1) Calculate the output \textit{Leader} of each classifier and choose \textit{Leader} classifier;
 \STATE (2) Reconstruct \textit{Leader}'s label as soft pseudo-label;
 \FOR{ Each AC}
 \STATE Determine whether to update using DLFM (Eq.~\ref{DLFM});
 \IF{needs to be updated}
 \STATE Update parameters through Eq.~\ref{eq11};
 \STATE Update memory bank through Eq.~\ref{eq12}.
 \ENDIF
 \ENDFOR
 \ENDFOR
	\end{algorithmic} 
\end{algorithm}

\section{Benchmarks,Backbone and Implementation Details}\label{D}
\subsection{Details about the Benchmarks} 
ALL experiments are conducted on the two popular multi-modal datasets Kinetics \citep{kay2017kinetics} and VGGSound \citep{chen2020vggsound}. \cite{yang2024test} further provides their corrupted visual and audio modality for TTA tasks.

Kinetics50 is a subset of the Kinetics dataset \citep{kay2017kinetics}, consisting of 50 randomly selected classes \citep{yang2024test}. It primarily includes videos that focus on human motion-related activities, with each clip lasting approximately 10 seconds and labeled with a single action class. All videos are sourced from YouTube. The Kinetics50 dataset comprises 29,204 visual-audio pairs for training and 2,466 pairs for testing, with the video modality playing a more prominent role in modality pairing.

VGGSound is a large-scale audio-visual dataset containing short audio clips extracted from YouTube videos \citep{chen2020vggsound}, covering 309 distinct everyday audio events. Each clip has a fixed duration of 10 seconds. For this study, we utilize the 157,602 pairs for training and 14,046 pairs for testing. Compared to Kinetics50, VGGSound includes a wider range of categories, introducing additional complexity to the classification task.

\begin{figure}
	\centering
	\includegraphics[width=0.85\linewidth]{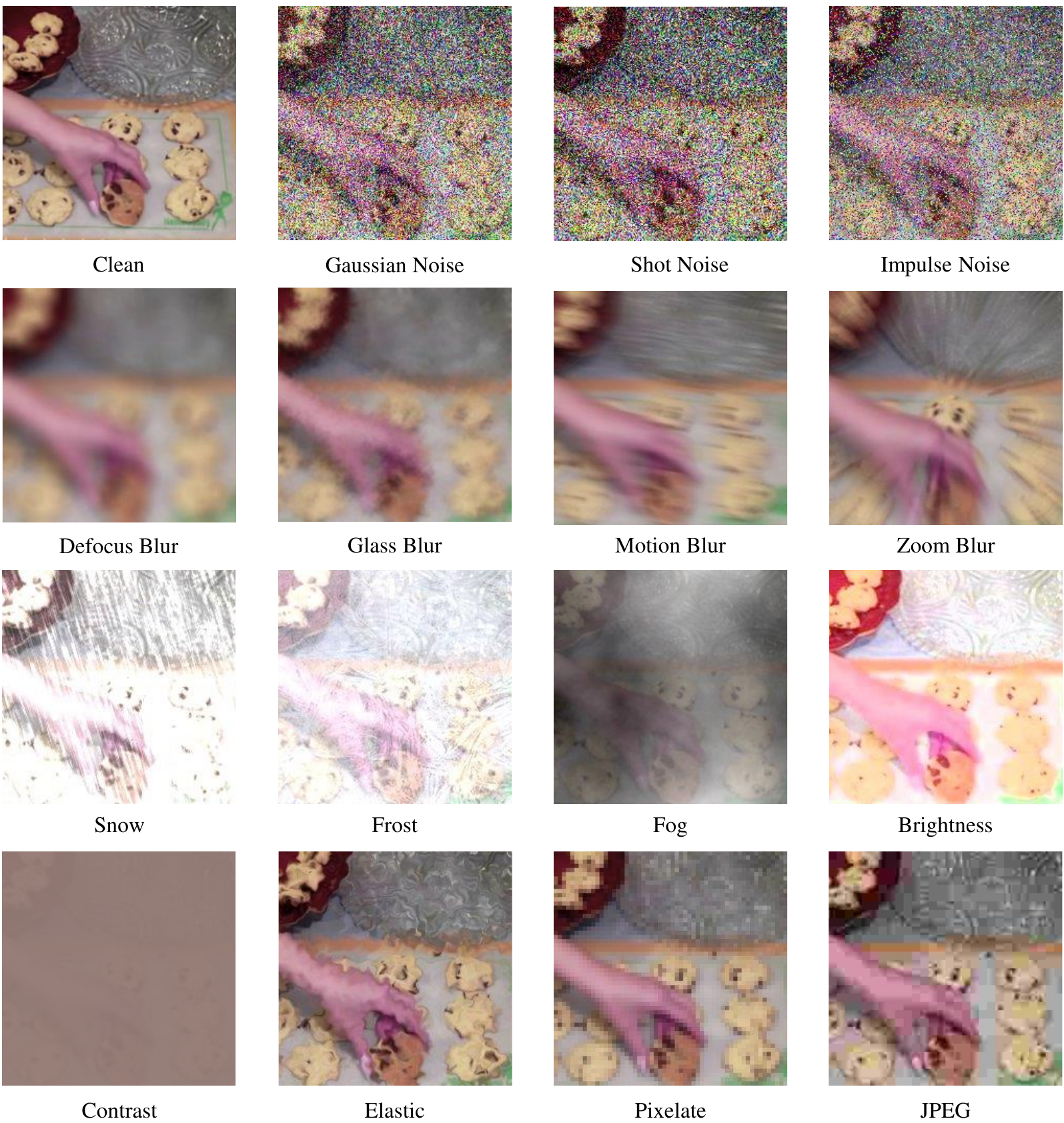}
	\caption{Visualization of 15 corruption types on the sampled video in Kinetics50-C benchmark.}
        \Description{}
	\label{fig.5}
\end{figure}

\begin{figure}
	\centering
	\includegraphics[width=0.9\linewidth]{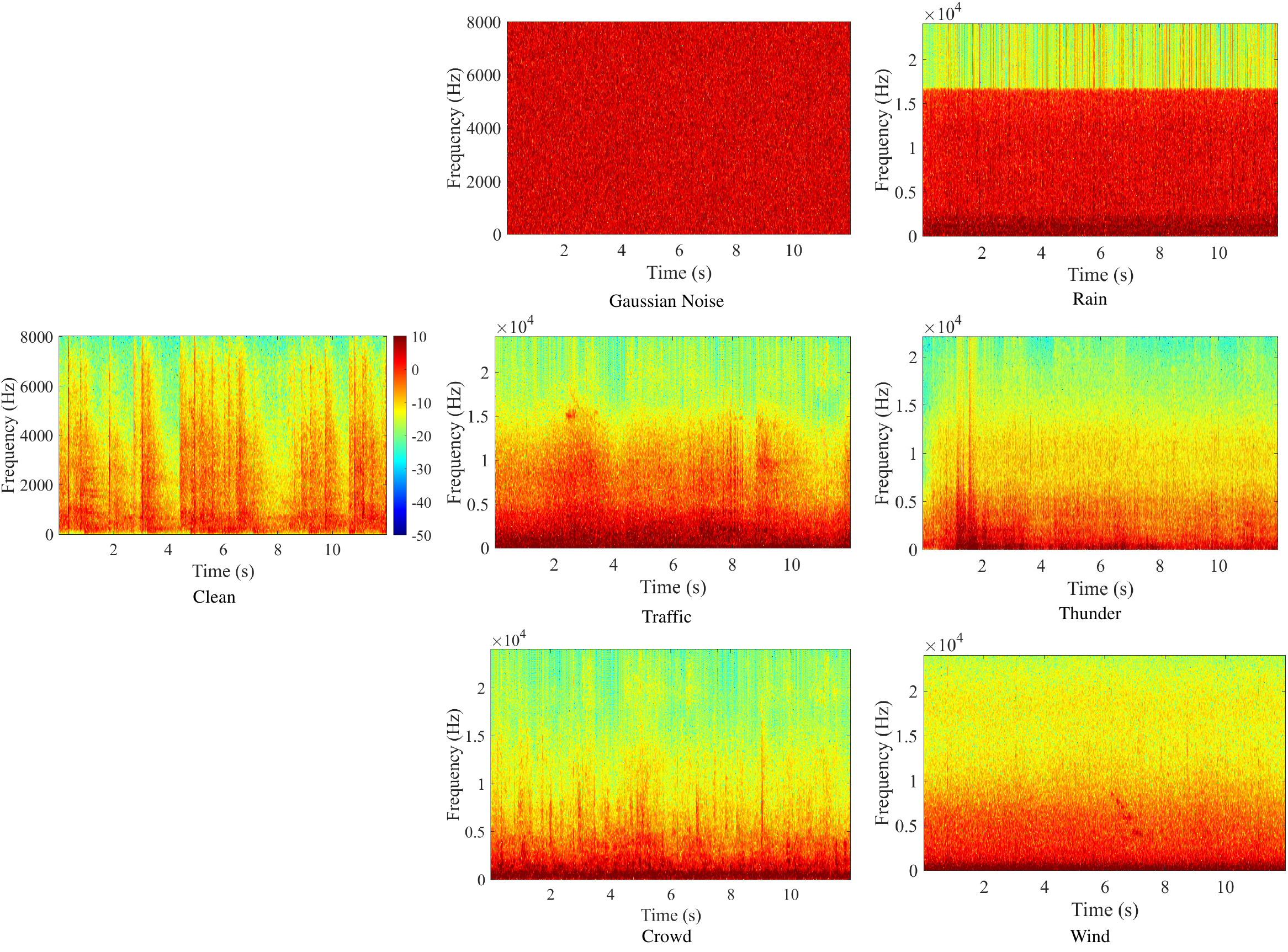}
	\caption{Spectrogram visualization of the clean audio and the corresponding 6 corruption
 types on the constructed Kinetics50-C benchmark. All Spectrogram use the same range of colorbar.}
        \Description{}
	\label{fig.6}
\end{figure}

Both datasets’ visual and audio modalities were extracted following the method described in \cite{gong2022contrastive}. To systematically explore the distributional shifts across modalities, various corruption types were applied to both visual and audio components. Following \cite{yang2024test}, 15 corruption types were introduced for the visual modality, each with five levels of severity for comprehensive evaluation. These corruptions include ``gaussian nois'', ``shot noise", ``impulse noise", ``defocus blur", ``glass blur", ``motion blur", ``zoom blur", ``snow", ``frost", ``fog", ``brightness", ``contrast", ``elastic transform", ``pixelate", and ``jpeg compression". Similarly, the audio modality was subjected to 6 different corruptions: ``gaussian noise", ``traffic noise", ``crowd noise", ``rain", ``thunder", and ``wind". The corrupted versions of these benchmarks are referred to as Kinetics50-C and VGGSound-C, respectively. Visualizations of sample corrupted video frames and audio spectrograms are provided in Figure~\ref{fig.5} and Figure~\ref{fig.6}.

\subsection{CAV-MAE Backbone}
CAV-MAE is employed as the pre-trained models for MM-CTTA in this paper. Its architecture consists of 11 Transformer blocks (known as feature encoder networks) dedicated to modality-specific feature extraction, followed by an additional Transformer block (known as fusion network) responsible for cross modal fusion. For the video input, 10 frames are sampled from each clip, from which a single frame is randomly selected and fed into the Transformer encoder for the visual modality. In the case of the audio input, the original 10-second audio waveform is transformed into a 2D spectrogram before being processed by the Transformer encoder for the audio modality \citep{gong2022note}.

\begin{table}[htp]
    \caption{Ablation studies on parameter $\gamma$.}
    \resizebox{0.3\textwidth}{!}{
    \begin{tabular}{|l|cc|cc|}
    \hline 
          \multirow{2}*{$\gamma$}&\multicolumn{2}{c|}{KS-video}&\multicolumn{2}{c|}{VGG-audio}  \\ \cline{2-5}
        &$\xrightarrow{}$&$\xleftarrow{}$&$\xrightarrow{}$&$\xleftarrow{}$\\ \hline
          1e-3&2.10&2.08&0.51&0.47  \\
          1e-2&2.03&2.12&0.31&0.39 \\
          1e-1&64.87&69.03&0.32&0.37 \\
          1e0&65.43&69.30&0.36&0.40  \\
          1e1&65.27&69.23&35.82&35.77\\
          1e2&65.29&68.95&35.78&35.76\\
          1e3&59.32&59.94&30.36&30.52\\\hline  
    \end{tabular}
    \label{gamma1}}
\end{table}

\begin{table*}[ht]
\caption{Comparison with SOTA methods on \textbf{audio} single-modality continual corruption task in terms of classification Top-1 accuracy (\%), using dataset \textbf{Kinetics50-C} and \textbf{VGGSound-C} in severity level 5. The best results for each domain are highlighted in \textbf{bold}. $^*$ means we revise the method from BN to LN for fair comparison.}
	\centering
	\resizebox{0.95\textwidth}{!}{
        \renewcommand\arraystretch{1}
		\begin{tabular}{|c|c|cccccc|c|cccccc|c|}
			\hline 
			\multirow{2}*{Method}&\multirow{2}*{Type}&\multicolumn{7}{c|}{Kinetics50-C}&\multicolumn{7}{c|}{VGGSound-C}\\ 
                \cline{3-16}
			&& Gauss. & Traff. & Crowd & Rain & Thund. & Wind & Avg. & Gauss. & Traff. & Crowd & Rain & Thund. & Wind & Avg. \\ \hline 
			&&\multicolumn{6}{c|}{$t\quad\xrightarrow[\quad\quad\quad\quad\quad\quad\quad\quad\quad\quad\quad\quad\quad\quad]{}$}&&\multicolumn{6}{c|}{$t\quad\xrightarrow[\quad\quad\quad\quad\quad\quad\quad\quad\quad\quad\quad\quad\quad\quad]{}$}& \\ 
                \hline 
                Source& - &73.97 & 65.21 & 67.79 & 70.27 & 67.98 & 70.45 & 69.28 & 37.32 & 21.24 & 16.89 & 21.82 & 27.37 & 25.66 & 25.05 \\
                TENT*& TTA &74.44 & 68.04 & 71.30 & 70.25 & 72.53 & 70.35 & 71.15 & 10.76 & 1.15 & 0.40 & 0.32 & 0.51 & 0.31 & 2.24 \\
                SAR& TTA &73.88 & 65.68 & 68.00 & 70.91 & 69.07 & 70.45 & 69.66 & 37.39 & 8.57 & 7.03 & 12.58 & 10.77 & 13.71 & 15.01 \\
                EATA& CTTA &73.95 & 65.26 & 68.03 & 70.45 & 68.17 & 70.48 & 69.39 & 40.49 & 31.07 & 31.98 & 31.40 & 38.26 & 33.84 & 34.51 \\
                MMTTA*& MM-TTA &69.32 & 69.01 & 69.00 & 69.07 & 68.96 & 68.95 & 69.05 & 14.40 & 1.92 & 0.84 & 0.36 & 0.47 & 0.31 & 3.05 \\
                READ& MM-TTA &74.74 & 68.88 & 70.43 & 70.69 & 72.31 & 69.73 & 71.13 & 40.51 & 25.39 & 20.38 & 20.06 & 21.14 & 16.07 & 23.93 \\
                \rowcolor{lightgray!45}
                MDAA& MM-CTTA &73.33 & 71.99 & 73.36 & 73.26 & 74.24 & 73.76 & 73.32 &38.57 &34.57 	&34.37 &34.30 &37.40 &35.70 &35.82 \\\hline
                &&\multicolumn{6}{c|}{$t\quad\xleftarrow[\quad\quad\quad\quad\quad\quad\quad\quad\quad\quad\quad\quad\quad\quad]{}$}&&\multicolumn{6}{c|}{$t\quad\xleftarrow[\quad\quad\quad\quad\quad\quad\quad\quad\quad\quad\quad\quad\quad\quad]{}$} &\\ 
                \hline 
                Source& - &74.00 & 65.20 & 67.92 & 70.24 & 68.01 & 70.43 & 69.30 & 37.31 & 21.25 & 16.89 & 21.81 & 27.37 & 25.66 & 25.05 \\
                TENT*& TTA &73.31 & 70.00 & 71.57 & 70.30 & 68.87 & 71.09 & 70.86 & 0.37 & 0.30 & 0.30 & 0.31 & 0.97 & 3.73 & 1.00 \\
                SAR& TTA &73.29 & 66.79 & 68.14 & 70.71 & 67.97 & 70.36 & 69.55 & 32.02 & 9.92 & 7.04 & 9.49 & 11.41 & 16.03 & 14.32 \\
                EATA& CTTA &73.96 & 65.30 & 68.15 & 70.36 & 68.18 & 70.40 & 69.39 & 40.65 & 32.32 & 30.65 & 32.23 & 38.16 & 32.39 & 34.40 \\
                MMTTA*& MM-TTA &68.92 & 69.47 & 70.50 & 69.59 & 69.62 & 69.63 & 69.62 & 0.35 & 0.32 & 0.30 & 0.23 & 0.67 & 3.25 & 0.85 \\
                READ& MM-TTA &73.11 & 70.16 & 69.68 & 70.90 & 72.07 & 70.73 & 71.11 & 24.23 & 16.23 & 16.84 & 17.65 & 29.12 & 30.30 & 22.39 \\
                \rowcolor{lightgray!45}
                MDAA& MM-CTTA &72.55 & 73.67 & 72.98 & 73.04 & 73.18 & 71.55 & 72.83 &38.59&35.35&34.50 &34.31&36.94&34.91&35.77 \\ \hline 
		\end{tabular}
}
\end{table*}

\subsection{Implementation Details}
In the final step of Algorithm \ref{alg1}, we determine the hyperparameters for MDAA. The expansion layer dimension, denoted as $\varphi$, theoretically benefits from larger values. However, an excessively large dimension may introduce a significant number of parameters, occupying considerable computational resources. Given the constraints of our available GPU resources, we set the dimension of $\varphi$ to 8000. The necessity of the parameter $\gamma$ in Eq.~\ref{eq5} has been established in \cite{zhuang2022acil}. The model demonstrates stable performance over a wide range of $\gamma$ values, indicating that as long as $\gamma$ is within a reasonable range, its impact on the model’s performance remains minimal. Table~\ref{gamma1} shows the results under a sweep over six orders of magnitude (\textit{i.e.,} ${10^{-3}, 10^{-2}, \dots, 10^{3}}$) using the same setting in the ablation study. The model performs robustly with a sufficient regularization, while collapses when the  value of $\gamma$ turns to be too extreme (larger than $10^{3}$ and smaller than $10^{-2}$).
In this paper, we set $\gamma$ to 1 for Kinetics-50 and 10 for VGGSound. As discussed in Sections \ref{4.2} and \ref{4.3}, the threshold $\theta$ in DLFM is fixed at 0.001 for both datasets, while the parameter $n$ for reconstructing soft labels is set to 2 for Kinetics-50 and 7 for VGGSound. All experiments were conducted on an RTX3090 GPU, with results averaged over three runs.


\section{More Experiment Results} \label{E}
In this appendix, we provide four more experiments for reference where Appendix~\ref{E.1} and \ref{E.2} are the supplementary comparative studies while Appendix~\ref{E.3} is the further discussion on hyperparameters applied in MDAA. 

\begin{table*}[ht]
\caption{Comparison with SOTA methods on \textbf{video} single-modality continual corruption task in terms of classification Top-1 accuracy (\%), with dataset \textbf{Kinetics50-C} in severity level 5.}
	\centering
	\resizebox{1\textwidth}{!}{
        \renewcommand\arraystretch{1}
		\begin{tabular}{|c|c|ccccccccccccccc|c|}
			\hline 
			Method&Type& Gauss. & Shot. & Impul. & Defoc. & Glass. & Motion. & Zoom. & Snow & Frost & Fog & Bright. & Cont.& Elastic. &Pixel. &Jpeg & Avg. \\ 
                \hline 
			&&\multicolumn{15}{c|}{$t\quad\xrightarrow[\qquad\qquad\qquad\qquad\qquad\qquad\qquad\qquad\qquad\qquad\qquad\qquad\qquad\qquad]{}$}& \\ 
                \hline 
                Source &-&48.67 & 49.81 & 49.01 & 67.77 & 61.88 & 70.95 & 66.19 & 61.39 & 61.42 & 45.35 & 75.94 & 51.86 & 65.81 & 68.77 & 66.10 & 60.73 \\
                TENT*&TTA&48.53 & 48.65 & 46.06 & 62.91 & 62.15 & 65.77 & 63.83 & 54.78 & 54.64 & 33.79 & 36.79 & 19.86 & 11.75 & 3.82 & 3.38 & 41.11 \\
                SAR&TTA&48.48 & 49.87 & 48.71 & 66.92 & 62.56 & 70.58 & 66.77 & 59.25 & 60.50 & 47.19 & 75.34 & 50.77 & 65.23 & 66.91 & 64.01 & 60.21 \\
                EATA&CTTA&48.76 & 49.84 & 49.03 & 67.78 & 62.02 & 70.92 & 66.20 & 61.55 & 61.53 & 45.38 & 75.97 & 51.90 & 65.91 & 68.74 & 66.09 & 60.77 \\
                MMTTA*&MM-TTA&48.74 & 49.05 & 48.88 & 49.12 & 48.94 & 48.88 & 48.86 & 48.92 & 48.92 & 48.86 & 49.03 & 48.79 & 48.88 & 48.80 & 48.92 & 48.91 \\
                READ&MM-TTA&51.02 & 53.54 & 54.24 & 68.16 & 66.36 & 68.95 & 67.76 & 62.85 & 64.72 & 59.62 & 70.99 & 53.02 & 67.03 & 63.71 & 62.78 & 62.32 \\
                \rowcolor{lightgray!45}
                MDAA&MM-CTTA&55.84 & 55.79 & 55.50 & 64.29 & 63.71 & 68.04 & 68.10 & 66.06 & 68.01 & 65.71 & 72.97 & 66.36 & 70.04 & 70.51 & 70.53 & 65.43 \\
                \hline
                &&\multicolumn{15}{c|}{$t\quad\xleftarrow[\qquad\qquad\qquad\qquad\qquad\qquad\qquad\qquad\qquad\qquad\qquad\qquad\qquad\qquad]{}$}& \\ 
                \hline
                Source &-&48.68 & 49.79 & 48.97 & 67.69 & 61.82 & 70.93 & 66.18 & 61.39 & 61.37 & 45.29 & 75.99 & 51.89 & 65.78 & 68.71 & 66.14 & 60.71 \\
                TENT*&TTA&41.93 & 47.19 & 49.22 & 64.80 & 64.33 & 68.05 & 63.86 & 61.42 & 62.12 & 51.65 & 73.92 & 50.80 & 67.47 & 69.03 & 66.80 & 60.17 \\
                SAR&TTA&48.08 & 49.07 & 48.83 & 66.25 & 63.02 & 70.39 & 66.27 & 59.53 & 60.31 & 46.28 & 75.62 & 51.23 & 65.47 & 68.47 & 66.14 & 60.33 \\
                EATA&CTTA&48.67 & 49.80 & 49.16 & 67.68 & 62.02 & 70.92 & 66.13 & 61.57 & 61.46 & 45.17 & 76.02 & 51.95 & 65.90 & 68.71 & 66.22 & 60.76 \\
                MMTTA*&MM-TTA&49.82 & 49.90 & 49.85 & 49.94 & 49.86 & 49.92 & 49.89 & 49.97 & 49.94 & 49.79 & 50.54 & 50.01 & 54.33 & 56.20 & 59.99 & 51.33 \\
                READ&MM-TTA&52.00 & 52.28 & 51.02 & 62.43 & 64.47 & 66.04 & 65.38 & 61.39 & 65.10 & 62.96 & 75.14 & 53.59 & 68.89 & 69.95 & 68.20 & 62.59 \\
                \rowcolor{lightgray!45}
                MDAA&MM-CTTA&70.55 & 70.43 & 70.08 & 71.81 & 70.30 & 72.19 & 70.88 & 69.38 & 69.35 & 67.48 & 73.16 & 62.07 & 68.63 & 67.98 & 65.21 & 69.30 \\ \hline 
		\end{tabular}
}
\end{table*}
\begin{table*}[ht]
\caption{Comparison with SOTA methods on \textbf{video} single-modality continual corruption task in terms of classification Top-1 accuracy (\%), with dataset \textbf{VGGSound-C} in severity level 5.}
	\centering
	\resizebox{1\textwidth}{!}{
        \renewcommand\arraystretch{1}
		\begin{tabular}{|c|c|ccccccccccccccc|c|}
			\hline 
			Method&Type& Gauss. & Shot. & Impul. & Defoc. & Glass. & Motion. & Zoom. & Snow & Frost & Fog & Bright. & Cont.& Elastic. &Pixel. &Jpeg & Avg. \\ 
                \hline 
			&&\multicolumn{15}{c|}{$t\quad\xrightarrow[\qquad\qquad\qquad\qquad\qquad\qquad\qquad\qquad\qquad\qquad\qquad\qquad\qquad\qquad]{}$}& \\ 
                \hline 
                Source &-&53.05 & 52.91 & 52.98 & 57.19 & 57.37 & 58.37 & 57.45 & 56.37 & 56.47 & 55.41 & 59.19 & 53.75 & 57.19 & 56.44 & 57.33 & 56.10 \\
                TENT*&TTA&53.27 & 52.76 & 52.00 & 54.58 & 54.35 & 55.04 & 54.86 & 52.59 & 52.81 & 53.11 & 53.50 & 50.80 & 53.03 & 52.31 & 52.14 & 53.14 \\
                SAR&TTA&53.14 & 53.29 & 53.21 & 56.95 & 56.91 & 58.41 & 57.47 & 56.10 & 56.78 & 56.34 & 58.35 & 53.98 & 57.27 & 55.93 & 56.48 & 56.04 \\
                EATA&CTTA&53.74 & 53.67 & 53.68 & 57.20 & 57.26 & 58.53 & 57.93 & 56.38 & 56.67 & 56.23 & 59.04 & 53.63 & 58.19 & 57.36 & 57.48 & 56.47 \\
                READ&MM-TTA&53.77 & 54.26 & 54.28 & 58.04 & 58.00 & 59.09 & 58.84 & 57.43 & 58.18 & 58.12 & 59.38 & 55.99 & 58.30 & 57.51 & 57.91 & 57.27 \\
                \rowcolor{lightgray!45}
                MDAA&MM-CTTA&55.63 & 55.91 & 55.88 & 57.50 & 57.77 & 58.27 & 57.84 & 57.05 & 56.72 & 56.77 & 58.67 & 55.78 & 57.55 & 56.87 & 57.84 & 57.07 \\
                \hline
                &&\multicolumn{15}{c|}{$t\quad\xleftarrow[\qquad\qquad\qquad\qquad\qquad\qquad\qquad\qquad\qquad\qquad\qquad\qquad\qquad\qquad]{}$}& \\ 
                \hline
                Source &-&53.01 & 52.88 & 52.97 & 57.20 & 57.42 & 58.41 & 57.49 & 56.37 & 56.49 & 55.43 & 59.16 & 53.74 & 57.19 & 56.42 & 57.29 & 56.10 \\
                TENT*&TTA&50.48 & 50.50 & 50.50 & 53.09 & 53.19 & 53.70 & 53.18 & 52.31 & 53.47 & 53.57 & 55.40 & 52.22 & 56.23 & 56.74 & 56.99 & 53.44 \\
                SAR&TTA&53.02 & 52.95 & 52.71 & 56.78 & 56.63 & 58.18 & 57.38 & 55.97 & 56.69 & 56.45 & 58.46 & 54.03 & 57.41 & 56.53 & 57.16 & 56.02 \\
                EATA&CTTA&53.69 & 53.66 & 53.61 & 57.26 & 57.34 & 58.46 & 57.84 & 56.35 & 56.82 & 56.72 & 59.21 & 54.10 & 58.28 & 57.39 & 57.54 & 56.55 \\
                READ&MM-TTA&55.33 & 55.34 & 54.93 & 58.16 & 57.88 & 58.82 & 58.50 & 57.24 & 57.94 & 57.86 & 59.65 & 55.67 & 58.79 & 57.87 & 58.10 & 57.47 \\
                \rowcolor{lightgray!45}
                MDAA&MM-CTTA&55.85 & 55.87 & 55.94 & 57.40 & 57.85 & 58.25 & 57.95 & 57.02 & 56.89 & 56.68 & 58.66 & 55.84 & 57.58 & 56.92 & 58.00 & 57.11 \\ \hline 
		\end{tabular}
}
\end{table*}

\subsection{Performance Comparison on Multi-modal Corruption} \label{E.1}
We follow the experimental setup in \citep{yang2024test} to examine whether our model remains superior to others when the adaptation task is not continuous. In this section, the corrupted target domains are independent, and the results for corrupted audio and video modalities are presented in Tables~9, 10 and 11. The results show that the SOTA performance of MDAA is due not only to its ability to combat catastrophic forgetting, but also to its strong capacity to handle MM-TTA tasks effectively.

\begin{table*}[H]
\caption{Comparison with SOTA methods on audio single-modality corruption task in terms of classification Top-1 accuracy (\%), using dataset \textbf{Kinetics50-C} and \textbf{VGGSound-C} in severity level 5. Results of the comparison methods are cite from \cite{yang2024test}.}\label{tableE1}
	\centering
	\resizebox{0.95\textwidth}{!}{
        \renewcommand\arraystretch{1}
		\begin{tabular}{|c|c|cccccc|c|cccccc|c|}
			\hline 
			\multirow{2}*{Method}&\multirow{2}*{Type}&\multicolumn{7}{c|}{Kinetics50-C}&\multicolumn{7}{c|}{VGGSound-C}\\ 
                \cline{3-16}
			&& Gauss. & Traff. & Crowd & Rain & Thund. & Wind & Avg. & Gauss. & Traff. & Crowd & Rain & Thund. & Wind & Avg. \\ \hline 
                Source& - & 73.7 &65.5& 67.9& 70.3& 67.9& 70.3& 69.3& 37.0 &25.5 &16.8& 21.6 &27.3&25.5& 25.6 \\
                TENT& TTA & 73.9& 67.4& 69.2& 70.4 &66.5 &70.5& 69.6& 10.6& 2.6& 1.8& 2.8& 5.3& 4.1 &4.5 \\
                SAR& TTA & 73.7& 65.4 &68.2& 69.9& 67.2 &70.2 &69.1 &37.4& 9.5& 11.0 &12.1 &26.8 &23.7 &20.1 \\
                EATA& CTTA & 73.7& 66.1 &68.5 &70.3 &67.9 &70.1 &69.4 &39.2& 26.1 &22.9& 26.0& 31.7& 30.4& 29.4 \\
                MMTTA& MM-TTA & 70.8 &69.2& 68.5& 69.0 &69.8 &69.4 &69.4 &14.1 &5.2 &6.4 &6.9& 8.6& 4.5 &7.6 \\
                READ& MM-TTA & 74.1& 69.0& 69.7 &71.1& 71.8 &70.7 &71.1& 40.4 &28.9 &26.6 &30.9 &36.7 &30.6 &32.4 \\
                \rowcolor{lightgray!45}
                MDAA& MM-CTTA &73.8 & 70.3 & 71.0 & 70.9 & 72.8 & 71.4 & 71.7 & 38.6 & 34.9 & 34.6 & 34.3 & 37.4 & 35.2 & 35.8 \\\hline
		\end{tabular}
}
\end{table*}

\begin{table*}[H]
\caption{Comparison with SOTA methods on \textbf{video} single-modality corruption task in terms of classification Top-1 accuracy (\%), with dataset \textbf{Kinetics50-C} in severity level 5. Results of the comparison methods are cite from \cite{yang2024test}.}
	\centering
	\resizebox{0.95\textwidth}{!}{
        \renewcommand\arraystretch{1}
		\begin{tabular}{|c|c|ccccccccccccccc|c|}
			\hline 
			Method&Type& Gauss. & Shot. & Impul. & Defoc. & Glass. & Motion. & Zoom. & Snow & Frost & Fog & Bright. & Cont.& Elastic. &Pixel. &Jpeg & Avg. \\ 
                \hline 
                Source &-& 46.8& 48.0 &46.9 &67.5& 62.2& 70.8& 66.7 &61.6& 60.3&46.7&75.2 &52.1& 65.7&66.5&61.9& 59.9 \\
                TENT&TTA& 46.3 &47.0 &46.3 &67.2 &62.5 &71.0& 67.6& 63.1 &61.1&34.9&75.4 &51.6 &66.8&67.2&62.7& 59.4 \\
                SAR&TTA& 46.7 &47.4 &46.8& 67.0& 61.9 &70.4 &66.4& 61.8& 60.6&46.0&75.2& 52.1& 65.7&66.4&62.0 &59.8 \\
                EATA&CTTA& 46.8 &47.6 &47.1& 67.2 &62.7& 70.6 &67.2 &62.3 &60.9&46.7&75.2 &52.4 &65.9&66.8&62.5& 60.1 \\
                MMTTA&MM-TTA& 46.2 &46.6 &46.1 &58.8 &55.7 &62.6 &58.7 &52.6 &54.4&48.5&69.1 &49.3& 57.6&56.4&54.6& 54.5 \\
                READ&MM-TTA& 49.4 &49.7& 49.0 &68.0 &65.1 &71.2& 69.0 &64.5 &64.4&57.4&75.5& 53.6& 68.3&68.0&65.1& 62.5 \\
                \rowcolor{lightgray!45}
                MDAA&MM-CTTA&55.1 & 55.3 & 55.7 & 64.5 & 62.3 & 67.7 & 65.0 & 61.6 & 63.6 & 57.9 & 72.2 & 54.8 & 66.6 & 67.0 & 65.2 & 62.3 \\
               \hline 
		\end{tabular}
}
\end{table*}\label{E2}
\begin{table*}[H]
\caption{Comparison with SOTA methods on \textbf{video} single-modality corruption task in terms of classification Top-1 accuracy (\%), with dataset \textbf{VGGSound-C} in severity level 5. Results of the comparison methods are cite from \cite{yang2024test}.}
	\centering
	\resizebox{0.95\textwidth}{!}{
        \renewcommand\arraystretch{1}
		\begin{tabular}{|c|c|ccccccccccccccc|c|}
			\hline 
			Method&Type& Gauss. & Shot. & Impul. & Defoc. & Glass. & Motion. & Zoom. & Snow & Frost & Fog & Bright. & Cont.& Elastic. &Pixel. &Jpeg & Avg. \\ 
                \hline 
                Source &-& 52.8&52.7&52.7&57.2&57.2&58.7&57.6&56.4&56.6&55.6&58.9& 53.7& 56.9&55.8&56.9& 56.0 \\
                TENT&TTA& 52.7&52.7 &52.7 &56.7& 56.5& 57.9 &57.2& 55.9 &56.3&56.3&58.4 &54.0& 57.4&56.2&56.7& 55.8 \\
                SAR&TTA& 52.9&52.8 &52.9& 57.2 &57.1& 58.6 &57.6& 56.3 &56.7&55.9&58.9 &54.0& 57.0&56.0&57.0 &56.1 \\
                EATA&CTTA&53.0&52.8&53.0& 57.2& 57.1 &58.6& 57.8 &56.3 &56.8&56.4&59.0& 54.1& 57.4&56.1&57.0 &56.2 \\
                MMTTA&MM-TTA& 7.1 &7.3 &7.3 &44.8& 41.5& 48.0 &45.5 &27.4& 23.5&30.5&46.9& 24.2& 40.3&40.7&45.7 &32.0 \\
                READ&MM-TTA&53.6&53.6& 53.5 &57.9& 57.7& 59.4& 58.8& 57.2& 57.8&55.0&59.9& 55.2& 58.6&57.1&57.9& 56.9 \\
                \rowcolor{lightgray!45}
                MDAA&MM-CTTA&\textbf{54.89} & \textbf{55.25} & \textbf{55.32} & 63.89 & \textbf{62.49} & 67.26 & 65.86 & \textbf{64.32} & \textbf{65.31} & \textbf{61.86} & 73.20 & \textbf{61.60} & \textbf{67.83} & \textbf{69.22} & \textbf{68.69} & \textbf{63.80} \\
               \hline 
		\end{tabular}
}
\label{E3}
\end{table*}

\subsection{Performance Comparison on Single-modality Continual Corruption} \label{E.2}
In this section, we compare the performance of each method under single-modality corruption. This task is similar to the \textbf{progressive single-modality corruption} task described in the main text, but here we use a batch size of 64.

\subsection{Dynamic threshold update} \label{E.3}
The threshold $\theta$ we used in DLFM is a fixed number. In this section we attempt to update $\theta$ in a dynamic way during the adaptation. We define the threshold $\theta_t^i$ for classifier $i$ in time $t$ as

\begin{align}
\theta_t^i = \begin{cases} \theta_{t-1}^i+\lambda \left ( d_t^i - d_{t-1}^i \right )
  & ,\text{if } t>1 \\\theta_{ini} 
  & ,\text{if } t=1
\end{cases} \quad ,i=\left \{ a,v,m \right \} 
\label{eq8} \\
d_t^{i} = \frac{{\textstyle \sum_{k=1}^{N_k}} \left ( max(P_k^{leader})-max(P_k^{i}) \right ) }{N_k},i=\left \{ a,v,m \right \},  \label{eq9}
\end{align}
where $\lambda$ is the learning rate, $\theta_{ini}$ is the initial threshold and $N_k$ is the batch size.$d_t^{i}$ is calculated to reflect the gap between \textit{Leader} and \textit{follower} $i$. The original intention is to adjust the threshold size according to $d_t$ changes to eliminate statistical bias across domains. However, as shown in Table~\ref{theta}, this approach achieves lower performance while requiring more variables to be memorized. Therefore, only the fixed threshold is used in the formal method.

\begin{table}
    \caption{Ablation studies on parameter $\lambda$.}
    \resizebox{0.8\columnwidth}{!}{
    \begin{tabular}{|l|cc|cc|}
    \hline 
          \multirow{2}*{$\lambda$}&\multicolumn{2}{c|}{KS-video}&\multicolumn{2}{c|}{VGG-audio}  \\ \cline{2-5}
          &$\xrightarrow{}$&$\xleftarrow{}$&$\xrightarrow{}$&$\xleftarrow{}$\\ \hline
          \rowcolor{lightgray!45}
          0&\textbf{63.55}&\textbf{69.54}&34.37&\textbf{34.35}  \\
          0.01&63.49&69.31&34.35&34.30 \\
          0.05&63.36&69.43&34.38&34.31 \\
          0.1&63.18&69.42&34.40&34.29  \\
          0.2&62.93&69.33&\textbf{34.41}&34.31\\\hline  
    \end{tabular}}\label{theta}
\end{table}



\onecolumn
\section{Proofs of Theorems} \label{A}
\textit{Proof of {\textbf{Theorem 1}}}.
known the optimal problem in \textit{Eqn.}\ref{eq10} can be further written as:
\begin{align}
 \notag
    &\underset{{\textbf{W}}_{\textup{Tg},t}}{\text{argmin}} {\textstyle \sum_{k=1}^{N_{\textup{Sr}}}} \omega_{k}   (\textbf{y}_{\textup{Sr},k} - {\textbf{x}}_{\textup{exf},k}{\textbf{W}}_{\textup{Tg},t})^\top(\textbf{y}_{\textup{Sr},k} - {\textbf{x}}_{\textup{exf},k}{\textbf{W}}_{\textup{Tg},t})+ \left ( \bar {\textbf{Y}}_{\textup{Tg},1:t} - {\textbf{X}}_{\textup{exf},1:t} {\textbf{W}}_{\textup{Tg},t} \right )^\top\left ( \bar {\textbf{Y}}_{\textup{Tg},1:t} - {\textbf{X}}_{\textup{exf},1:t} {\textbf{W}}_{\textup{Tg},t} \right )  +{\gamma} \textbf{W}_{_{\textup{Tg},t}}^\top\textbf{W}_{_{\textup{Tg},t}}\\
\notag
    =&\underset{{\textbf{W}}_{\textup{Tg},t}}{\text{argmin}} {\textstyle \sum_{k=1}^{N_{\textup{Sr}}}} \omega_{k} ({\textbf{W}}_{\textup{Tg},t}^\top{\textbf{x}}_{\textup{exf},k}^\top{\textbf{x}}_{\textup{exf},k}{\textbf{W}}_{\textup{Tg},t}-\textbf{y}_{\textup{Sr},k}^\top{\textbf{x}}_{\textup{exf},k}{\textbf{W}}_{\textup{Tg},t}-{\textbf{W}}_{\textup{Tg},t}^\top{\textbf{x}}_{\textup{exf},k}^\top\textbf{y}_{\textup{Sr},k} +\textbf{y}_{\textup{Sr},k}^\top\textbf{y}_{\textup{Sr},k}) + \\
\notag
    &({\textbf{W}}_{\textup{Tg},t}^\top{\textbf{X}}_{\textup{exf},1:t}^\top{\textbf{X}}_{\textup{exf},1:t}{\textbf{W}}_{\textup{Tg},t}-\bar {\textbf{Y}}_{\textup{Tg},1:t}^\top{\textbf{X}}_{\textup{exf},1:t}{\textbf{W}}_{\textup{Tg},t}-{\textbf{W}}_{\textup{Tg},t}^\top{\textbf{X}}_{\textup{exf},1:t}^\top\bar {\textbf{Y}}_{\textup{Tg},1:t} +\bar {\textbf{Y}}_{\textup{Tg},1:t}^\top\bar {\textbf{Y}}_{\textup{Tg},1:t})
    + {\gamma} \textbf{W}_{\textup{Tg},t}^\top\textbf{W}_{\textup{Tg},t}.
\end{align}

Note above equation as $L_{1}$, derive $L_{1}$ for ${\textbf{W}}_{\textup{Tg},1:t}$ as
\begin{align}
\notag
    \frac{\partial L_{1}}{\partial {\textbf{W}}_{\textup{Tg},t}} &=  2{\textstyle \sum_{k=1}^{N_{\textup{Sr}}}} \omega_{k}(\textbf{x}_{\textup{exf},k}^\top\textbf{x}_{\textup{exf},k}{\textbf{W}}_{\textup{Tg},t}-\textbf{x}_{\textup{exf},k}^\top\textbf{y}_{\textup{Sr},k})+ ({\textbf{X}}_{\textup{exf},1:t}^\top{\textbf{X}}_{\textup{exf},1:t}{\textbf{W}}_{\textup{Tg},t}-{\textbf{X}}_{\textup{exf},1:t}^\top\bar {\textbf{Y}}_{\textup{Tg},1:t})+2{\gamma} {\textbf{W}}_{\textup{Tg},t} \\
\notag
    & = 2{\textstyle \sum_{k=1}^{N_{\textup{Sr}}}} (\tilde{\textbf{x}}_{\textup{exf},k}^\top\tilde{\textbf{x}}_{\textup{exf},k}{\textbf{W}}_{\textup{Tg},t}-\tilde{\textbf{x}}_{\textup{exf},k}^\top\tilde{\textbf{y}}_{\textup{Sr},k})+ ({\textbf{X}}_{\textup{exf},1:t}^\top{\textbf{X}}_{\textup{exf},1:t}{\textbf{W}}_{\textup{Tg},t}-{\textbf{X}}_{\textup{exf},1:t}^\top\bar {\textbf{Y}}_{\textup{Tg},1:t}) +2{\gamma} {\textbf{W}}_{\textup{Tg},t}\\
\notag
    &=2(\tilde{\textbf{X}}_{\textup{exf},k}^\top\tilde{\textbf{X}}_{\textup{exf},k}{\textbf{W}}_{\textup{Tg},t} + \textbf{X}_{\textup{exf},1:t}^\top \textbf{X}_{\textup{exf},1:t}{\textbf{W}}_{\textup{Tg},t}-\tilde{\textbf{X}}_{\textup{exf},k}^\top\tilde{\textbf{Y}}_{\textup{Sr},k} -{\textbf{X}}_{\textup{exf},1:t}^\top{\bar{\textbf{Y}}}_{\textup{Tg},1:t})+ 2{\gamma} {\textbf{W}}_{\textup{Tg},t}\\
\notag
    &=0.
\end{align}

Therefore 
\begin{align}
\notag
\hat{\textbf{W}}_{\textup{Tg},t} = ( {\tilde{\textbf{X}} }_{\textup{exf},\textup{Sr}}^\top\tilde{\textbf{X}}_{\textup{exf},\textup{Sr}}+ \textbf{X}_{\textup{exf},1:t}^\top \textbf{X}_{\textup{exf},1:t}+\gamma{\textbf{I}})^{-1}({\tilde{\textbf{X}} }_{\textup{exf},\textup{Sr}}^\top{\tilde{\textbf{Y}}}_{\textup{Sr}}+{\textbf{X}}_{\textup{exf},1:t}^\top{\bar{\textbf{Y}}}_{\textup{Tg},1:t})=\textbf{P}_{\textup{Tg},t}^{-1}\textbf{Q}_{\textup{Tg},t},\\
\notag
\textbf{P}_{\textup{Tg},t}= \textbf{P}_{\textup{Sr}}+ \textbf{X}_{\textup{exf},1:t}^\top \textbf{X}_{\textup{exf},1:t}=\textbf{P}_{\textup{Tg},1}+ \textbf{X}_{\textup{exf},2:t}^\top \textbf{X}_{\textup{exf},2:t}
       = \dots =\textbf{P}_{\textup{Tg},t-1}+ \textbf{X}_{\textup{exf},t}^\top \textbf{X}_{\textup{exf},t},\\
\notag
  \textbf{Q}_{\textup{Tg},t}=\textbf{Q}_{\textup{Sr}}+{\textbf{X}}_{\textup{exf},1:t}^\top{\bar{\textbf{Y}}}_{\textup{Tg},1:t}=\textbf{Q}_{\textup{Tg},1}+{\textbf{X}}_{\textup{exf},2:t}^\top{\bar{\textbf{Y}}}_{\textup{Tg},2:t}
       = \dots =\textbf{Q}_{\textup{Tg},t-1}+ \textbf{X}_{\textup{exf},t}^\top {\bar{\textbf{Y}}}_{\textup{Tg},t}.
\end{align}

\end{document}